\documentclass[10pt,twocolumn,letterpaper]{article}

\usepackage{cvpr}
\usepackage{times}
\usepackage{epsfig}
\usepackage{graphicx}
\usepackage{amsmath}
\usepackage{amssymb}
\usepackage{amsthm}
\usepackage{eqnarray}
\usepackage{booktabs}

\newtheorem{theorem}{Theorem}
\numberwithin{theorem}{section}
\newtheorem{proposition}[theorem]{Proposition}

\newcommand{\C}{{\mathbb C}}

\newcommand{\M}[1]{\mathtt{#1}}
\newcommand{\V}[1]{\mathbf{#1}}

\newcommand{\comment}[1]{}

\def\Hk{Heikkil{\"{a}}\xspace}  
\def\gb{Gr{\"o}bner basis\xspace}
\def\gf{Gr{\"o}bner fan\xspace}
\def\GB{Gr{\"o}bner Basis\xspace}
\def\gbs{Gr{\"o}bner bases\xspace}

\usepackage[pagebackref=true,breaklinks=true,letterpaper=true,colorlinks,bookmarks=false]{hyperref}

\cvprfinalcopy 

\ifcvprfinal\fi
\begin{document}

\title{A sparse resultant based method for efficient minimal solvers}

\author{Snehal Bhayani\\
Center for Machine Vision and Signal Analysis\\
University of Oulu, Finland\\
{\tt\small snehal.bhayani@oulu.fi}
\and
Zuzana Kukelova\\
 Center for Machine Perception\\
Czech Technical University, Prague\\
{\tt\small kukelova@cmp.felk.cvut.cz}
\and
Janne \Hk\\
Center for Machine Vision and Signal Analysis\\
University of Oulu, Finland\\
{\tt\small janne.heikkila@oulu.fi}
}

\maketitle
\setlength{\textfloatsep}{10pt plus4pt minus1pt}
\begin{abstract}
\noindent Many computer vision applications require robust and efficient estimation of camera geometry. The robust estimation is usually based on solving camera geometry problems from a minimal number of input data measurements, i.e. solving minimal problems in a RANSAC framework. Minimal problems often result in complex systems of polynomial equations. Many state-of-the-art efficient polynomial solvers to these problems are based on \gbs and the action-matrix method that has been automatized and highly optimized in recent years.
In this paper we study an alternative algebraic method for solving systems of polynomial equations, i.e., the sparse resultant-based method and propose a novel approach to convert the resultant constraint to an eigenvalue problem. This technique can significantly improve the efficiency and stability of existing resultant-based solvers.
We applied our new resultant-based method to a large variety of computer vision problems and show that for most of the considered problems, the new method leads to solvers that are the same size as the the best available \gb solvers and of similar accuracy. For some problems the new sparse-resultant based method leads to even smaller and more stable solvers than the state-of-the-art \gb solvers. Our new method can be fully automatized and incorporated into existing tools for automatic generation of efficient polynomial solvers and as such it represents a competitive alternative to popular \gb methods for minimal problems in computer vision.
\end{abstract}

\section{Introduction}
\noindent Computing camera geometry is one of the most important tasks in computer vision~\cite{HZ-2003} with many applications e.g. in structure from motion~\cite{Snavely-IJCV-2008}, visual navigation~\cite{DBLP:journals/ram/ScaramuzzaF11}, large scale 3D reconstruction~\cite{DBLP:conf/cvpr/HeinlySDF15} and image localization~\cite{Sattler16PAMI}.

The robust estimation of camera geometry is usually based on solving so-called minimal problems~\cite{Nister-5pt-PAMI-2004,Kukelova-ECCV-2008, Kukelova-thesis}, \ie problems that are solved from minimal samples of input data, inside a RANSAC framework~\cite{Fischler-Bolles-ACM-1981,Chum-2003, DBLP:journals/pami/RaguramCPMF13}. Since the camera geometry estimation has to be performed many times in RANSAC~\cite{Fischler-Bolles-ACM-1981}, fast solvers to minimal problems are of high importance. Minimal problems often result in complex systems of polynomial equations in several variables. A popular approach for solving minimal problems 
is to design procedures that can efficiently solve only a special class of systems of equations, \eg systems resulting from the 5-pt relative pose problem~\cite{Nister-5pt-PAMI-2004}, and move as much computation as possible from the ``online'' stage of solving equations to an earlier pre-processing ``offline'' stage.

Most of the state-of-the-art specific minimal solvers are based on \gbs and the action-matrix method~\cite{Cox-Little-etal-05}. The \gb method was popularized in computer vision by Stewenius~\cite{DBLP:phd/basesearch/Stewenius05}. The first efficient \gb solvers were mostly handcrafted~\cite{Stewenius-ISPRS-2006,Stewenius-CVPR-2005} and sometimes very unstable~\cite{stewenius2005hard}. However, in the last 15 years much effort has been put into making the process of constructing the solvers more automatic~\cite{Kukelova-ECCV-2008,larsson2017efficient,larsson2017polynomial} and the solvers stable~\cite{byrod2007improving,byrod2008column} and more efficient~\cite{larsson2017efficient,larsson2017polynomial,larsson2016uncovering,Bujnak-CVPR-2012, DBLP:conf/cvpr/LarssonOAWKP18}. There are now powerful tools available for the automatic generation of efficient \gb solvers \cite{Kukelova-ECCV-2008,larsson2017efficient}.  

While the \gb method for generating efficient minimal solvers was deeply studied in computer vision and all recently generated \gb solvers are highly optimized in terms of efficiency and stability, little attention has been paid to an alternative algebraic method for solving systems of polynomial equations, \ie the resultant-based method. The resultant-based method was manually applied to several computer vision problems \cite{Kukelova-PolyEig-PAMI-2012, Hartley-PAMI-2012, Hartley-PAMI-2012, DBLP:conf/wacv/KastenGB19, Kukelova-thesis, Kukelova-PolyEig-PAMI-2012}. However in contrast to the \gb method, there is no general method for automatically generating efficient resultant-based minimal solvers. The most promising results in this direction were proposed by Emiris~\cite{emiris-general} and \Hk~\cite{DBLP:conf/iccv/Heikkila17}, where methods based on sparse resultants were proposed and applied to camera geometry problems. While these methods can be extended for general minimal problems that appear in computer vision and can be automatized, they usually lead (due to linearizations) to larger and less efficient solvers than \gb solvers.

In this paper, we propose a novel approach to generating minimal solvers using sparse resultants, which is based on adding an extra equation of a special form to the input system. Our algorithm is inspired by the ideas explored in~\cite{DBLP:conf/iccv/Heikkila17, emiris-general}, but thanks to the special form of added equation and by solving the resultant as a small eigenvalue problem, in contrast to a polynomial eigenvalue problem in~\cite{DBLP:conf/iccv/Heikkila17}, the new approach achieves significant improvements over~\cite{DBLP:conf/iccv/Heikkila17, emiris-general} in terms of efficiency of the generated solvers. 
Specifically our contributions include,
\vspace{-0.2cm} 
\begin{itemize}
\itemsep-0.3em
    \item A novel sparse resultant-based approach to generating polynomial solvers based on adding an extra equation of a special form and transforming the resultant matrix constraint to a regular eigenvalue problem.
    \item Two procedures to reduce the size of resultant matrix that lead to faster solvers than the best available state-of-the-art solvers for some minimal problems.
   \item A general method for automatic generation of efficient resultant-based polynomial solvers for many important minimal problems that achieves competitive performance in terms of speed and stability with respect to the best available state-of-the-art solvers generated by highly optimized \gb techniques~\cite{larsson2017efficient, DBLP:conf/cvpr/LarssonOAWKP18}. The automatic generator of resultant-based solvers will be made publicly available.
\end{itemize}
\section{Theoretical background and related work} 
\noindent In this paper we use notation and basic concepts from the book by Cox \etal~\cite{Cox-Little-etal-05}. Our objective is to solve $m$ polynomial equations, 
\begin{eqnarray}\label{eq:system}
\lbrace f_1(x_1,\dots,x_n)=0,\dots,f_m(x_1,\dots,x_n)=0 \rbrace
\end{eqnarray}
in $n$ unknowns, $X =\lbrace x_1,\dots,x_n \rbrace$. Let $\C[X]$ denote the set of all polynomials in unknowns $X$ with coefficients in $\C$. The ideal $I = \langle f_1,\dots,f_m \rangle \subset \C[X]$ is the set of all polynomial combinations of our generators $f_1,\dots,f_m$. The set $V$ of all solutions of the system~\eqref{eq:system}
is called the affine variety. Each polynomial $f \in I$ vanishes on the solutions of \eqref{eq:system}. Here we assume that the ideal $I$  generates a zero dimensional variety, \ie the system~\eqref{eq:system} has a finite number of solutions. Using the ideal $I$ we can define the quotient ring $A = \C[X]/I$ which is the set of equivalence classes over $\C[X]$ defined by the relation $a \sim b 
\iff (a - b) \in I$. 
If $I$ has a zero-dimensional variety
then the quotient ring $A = \C[X]/I$ is a finite-dimensional vector space over $\C$. For an ideal $I$ there exist special sets of generators called \gbs which have the nice property that the remainder after division is unique. Using a \gb we can define a linear basis for the quotient ring $A = \C[X]/I$.

\subsection{\GB method}\label{sec:GB}
\noindent \gbs can be used to solve our system of polynomial equations~\eqref{eq:system}.
One of the popular approaches for solving systems of equations using \gbs is the multiplication matrix method, known also as the action matrix method~\cite{Cox-Little-etal-05,Sturmfels-CBMS-2002}. This method was recently used to efficiently solve many of the minimal problems in computer vision~\cite{Kukelova-thesis, Kukelova-ECCV-2008,larsson2017efficient,DBLP:conf/cvpr/LarssonOAWKP18}. The goal of this method is to transform the problem of finding the solutions to~\eqref{eq:system} to a problem of eigendecomposition of a special multiplication matrix~\cite{Cox-IVA-2015}. Let us consider the mapping $T_f : A \rightarrow A$ of the multiplication by a polynomial $f \in \C[X]$. 
$T_f$ is a linear mapping for which $T_f = T_g$ iff $f - g \in I$. In our case $A$ is a finite-dimensional vector space over $\C$ and therefore we can represent $T_f$ 
by its matrix with respect to some linear basis $B$ of $A$. 
For a basis $B = ([b_1],\dots,[b_k])$ consisting of $k$ monomials, 
$T_f$ can be represented by $k \times k$ multiplication (action) matrix $\M{M}_{f} := (m_{ij})$ such that
$T_f ([b_j]) = [f b_j ] = \sum_{i=1}^k m_{ij}[b_i]$.
It can be shown~\cite{Cox-IVA-2015} that $\lambda \in \C$ is an eigenvalue of the matrix $\M{M}_{f}$ iff $\lambda$ is a value of the function $f$ on the variety $V$. In other words, if $f$ is e.g. $x_n$ then the eigenvalues of $\M{M}_{f}$ are the $x_n$-coordinates of the solutions of~\eqref{eq:system}. The solutions to the remaining variables can be obtained from the eigenvectors of $\M{M}_{f}$. This means that after finding the multiplication matrix $\M{M}_{f}$, we can recover the solutions by solving the eigendecompostion of $\M{M}_{f}$ for which efficient algorithms exist. Moreover, if the ideal $I$ is a radical ideal, \ie $I = \sqrt I$,~\cite{Cox-IVA-2015}, then $k$ is equal to the number of solutions to the system~\eqref{eq:system}. Therefore, \gb methods usually solve an eigenvalue problem of a size that is equivalent to the number of solutions of the problem. For more details and proofs we refer the reader to~\cite{Cox-Little-etal-05}. 

The coefficients of the multiplication matrix $\M{M}_{f}$ are polynomial combinations of coefficients of the input polynomials~\eqref{eq:system}. For computer vision problems these polynomial combinations are often found ``offline`` in a pre-processing step. In this step, a so-called \textit{elimination template} is generated, which is actually an expanded set of equations constructed by multiplying original equations with different monomials. This template matrix is constructed such that after filling it with coefficients from the input equations and performing Gauss-Jordan(G-J) elimination of this matrix, the coefficients of the multiplication matrix $\M{M}_{f}$ can be obtained from this eliminated template matrix.

The first automatic approach for generating elimination templates and \gb solvers was presented in~\cite{Kukelova-ECCV-2008}. Recently an improvement to the automatic generator~\cite{Kukelova-ECCV-2008} was proposed in~\cite{larsson2017efficient} to exploit the inherent relations between the input polynomial equations and it results in more efficient solvers than~\cite{Kukelova-ECCV-2008}. The automatic method from~\cite{larsson2017efficient} was later extended by a method for dealing with saturated ideals~\cite{larsson2017polynomial} and a method for detecting symmetries in polynomial systems~\cite{larsson2016uncovering}. 

In general, the answer to the question ``What is the smallest elimination template for a given problem?" is not known. In~\cite{DBLP:conf/cvpr/LarssonOAWKP18}  the authors showed that the method~\cite{larsson2017efficient}, which is based on the grevlex ordering of monomials and the so-called standard bases of the quotient ring $A$ is not optimal in terms of template sizes. The authors of~\cite{larsson2017efficient}  proposed two methods for generating smaller elimination templates. The first is based on enumerating and testing all \gbs w.r.t. different monomial orderings, i.e., the so-called \gf.
By generating solvers w.r.t. all these \gbs and using standard bases of the quotient ring $A$, smaller solvers were obtained for many problems.
The second method goes ``beyond \gbs'' and it uses a manually designed heuristic sampling scheme for generating ``non-standard" monomial bases $B$ of  $A = \C[X]/I$. This heuristic leads to more efficient solvers than the \gf method in many cases. While the \gf method will provably generate at least as efficient solvers as the grevlex-based method from~\cite{larsson2017efficient}, no proof can be in general given for the ``heuristic-based'' method. The proposed heuristic sampling scheme  
uses only empirical observations on which basis monomials will likely result in small templates and it samples a fixed number (1000 in the paper) of candidate bases consisting of these monomials. Even though, e.g. the standard grevlex monomial basis will most likely be sampled during the sampling, it is in general not clear how large templates it will generate for a particular problem. The results will also depend on the number of bases tested inside the heuristic. 


\subsection{Sparse Resultants} 
\noindent An alternate approach towards solving polynomial equations is that of using resultants. Simply put, a resultant is an irreducible polynomial constraining coefficients of a set of $n+1$ polynomials, $F = \lbrace f_{1}(x_1,\dots,x_n),\dots,f_{n+1}(x_1,\dots,x_n\rbrace)$ in $n$ variables to have a non-trivial solution. One can refer to Cox \etal~\cite{Cox-Little-etal-05} for a more formal theory on resultants. We have $n+1$ equations in $n$ variables because resultants were initially developed to determine whether a system of polynomial equations has a common root or not. If a coefficient of monomial $\V{x}^{\V{\alpha}}$ in the $i^{th}$ polynomial of $F$ is denoted as $u_{i,\V{\alpha}}$ the resultant is a polynomial $Res([u_{i,\V{\alpha}}])$ with $u_{i, \V{\alpha}}$ as variables.

Using this terminology, the basic idea for a resultant based method is to expand $F$ to a set of linearly independent polynomials which can be linearised as 
  $  \M{M}([u_{i, \V{\alpha}}]) \V{b},$
where $\V{b}$ is a vector of monomials of form $\V{x}^{\V{\alpha}}$ and $\M{M}([u_{i, \V{\alpha}}])$ has to be a square matrix that has full rank for generic values of $u_{i,\V{\alpha}}$, i.e. $\det \M{M}([u_{i, \V{\alpha}}]) \neq 0$. The determinant of the matrix $\M{M}([u_{i, \V{\alpha}}])$ is a non-trivial multiple of the resultant $Res([u_{i, \V{\alpha}}])$~\cite{Cox-Little-etal-05}. Thus $\det \M{M}([u_{i, \V{\alpha}}])$ must vanish, if the resultant vanishes, \ie $Res([u_{i, \V{\alpha}}]) = 0 \implies \det \M{M}([u_{i, \V{\alpha}}]) = 0$. It is known that $Res([u_{i, \V{\alpha}}])$ vanishes iff the polynomial system $F$ has a solution~\cite{Cox-Little-etal-05}. This gives us the necessary condition for the existence of roots of $F = 0$. Hence the equation $\det \M{M}([u_{i, \V{\alpha}}]) = 0$ gives us those values of $u_{i, \V{\alpha}}$ such that $F=0$ have a common root.

Resultants can be used to solve $n$ polynomial equations in $n$ unknowns. The most common approach used for this purpose is to \textit{hide a variable} by considering it as a constant. By hiding, say $x_{n}$, we obtain $n$ polynomials in $n\!\! -\!\! 1$ variables, so we can use the concept of resultants and compute $Res([u_{i, \V{\alpha}}], x_{n})$ which now becomes a function of $u_{i, \V{\alpha}}$ as well as $x_{n}$. Algorithms based on hiding a variable attempt to expand $F$ to a linearly independent set of polynomials that can be re-written in a matrix form as
\begin{equation} \label{eq:spresconstraint2}
    \M{M}([u_{i, m}],x_{n}) \V{b} = 0, 
\end{equation}
where $\M{M}([u_{i, \V{\alpha}}],x_{n})$ is a square matrix whose elements are polynomials in $x_{n}$ and coefficients $u_{i, \V{\alpha}}$ and $\V{b}$ is the vector of monomials in $x_{1},\dots,x_{n-1}$. For simplicity we will denote the matrix $\M{M}([u_{i, \V{\alpha}}],x_{n})$  as $\M{M}(x_{n})$ in the rest of this section. Here we actually estimate a multiple of the actual resultant via the determinant of the matrix $\M{M}(x_{n})$ in ~\eqref{eq:spresconstraint2}. This resultant is known as a \textit{hidden variable resultant} and it is a polynomial in $x_n$ whose roots are the $x_{n}$-coordinates of the solutions of the system of polynomial equations. For theoretical details and proofs see \cite{Cox-Little-etal-05}. Such a hidden variable approach has been used in the past to solve various minimal problems~\cite{Hartley-PAMI-2012, DBLP:conf/wacv/KastenGB19, Kukelova-thesis, Kukelova-PolyEig-PAMI-2012}. 

The most common way to solve the original system of polynomial equations is to transform~\eqref{eq:spresconstraint2} to a polynomial eigenvalue problem (PEP)~\cite{Cox-IVA-2015} that transforms~\eqref{eq:spresconstraint2} as
\begin{equation}\label{eq:pepformulation}
(\M{M}_{0} + \M{M}_{1} \ x_{n}+...+ \M{M}_{l} \ x^{l}_{n})\V{b} = \V{0},
\end{equation}
where $l$ is the degree of the matrix $\M{M}(x_{n})$ in the hidden variable $x_n$ and matrices $\M{M}_{0},...,\M{M}_{l}$ are matrices that depend only on the coefficients $u_{i, \V{\alpha}}$ of the original system of polynomials. The PEP~\eqref{eq:pepformulation} can be easily converted to a generalized eigenvalue problem (GEP):
\begin{equation}\label{eq:GEP}
\M{A} \V{y} =  x_{n} \M{B} \V{y},
\end{equation}
and solved using standard efficient eigenvalue algorithms. Basically, the eigenvalues give us the solution to $x_{n}$ and the rest of the variables can be solved from the corresponding eigenvectors, $\V{y}$~\cite{Cox-Little-etal-05}. But this transformation to a GEP relaxes the original problem of finding the solutions to our input system and computes eigenvectors that do not satisfy the monomial dependencies induced by the monomial vector $\V{b}$. And many times it also introduces extra parasitic (zero) eigenvalues leading to slower polynomial solvers.

Alternately, we can add a new polynomial 
\begin{equation}\label{eq:u-res}
f_{n+1} = u_0 + u_1 x_1 + \dots + u_n x_n
\end{equation} to $F$ and compute a so called \textit{$u$-resultant}~\cite{Cox-Little-etal-05} by hiding $u_0,\dots,u_n$. In general random values are assigned to $u_1,\dots,u_n$. The $u$-resultant matrix is computed from these $n+1$ polynomials in $n$ variables in a way similar to the one explored above. For more details about $u$-resultant one can refer to~\cite{Cox-Little-etal-05}.

For sparse polynomial systems it is possible to obtain more compact resultants using specialized algorithms. Such resultants are commonly referred to as \textit{Sparse Resultants}. A sparse resultant would mostly lead to a more compact matrix $\M{M}(x_{n})$ and hence a smaller eigendecomposition problem. Emiris \etal \cite{DBLP:conf/issac/EmirisC93, DBLP:journals/jacm/CannyE00} have proposed a generalised approach for computing sparse resultants using mixed-subdivision of polytopes.
Based on \cite{DBLP:conf/issac/EmirisC93, DBLP:journals/jacm/CannyE00} Emiris proposed a method for generating a resultant-based solver for sparse systems of polynomial equations, that was divided in ``offline'' and ``online'' computations. The resulting solvers were based either on the hidden-variable trick~\eqref{eq:spresconstraint2} or the u-resultant of the general form~\eqref{eq:u-res}. As such the resulting solvers were usually quite large and not very efficient.
More recently \Hk~\cite{DBLP:conf/iccv/Heikkila17} have proposed an improved approach to test and extract smaller $\M{M}(x_n)$. This method transforms~\eqref{eq:spresconstraint2} to a GEP~\eqref{eq:GEP} and solves for eigenvalues and eigenvectors to compute solutions to unknowns. The methods \cite{DBLP:journals/jacm/CannyE00,emiris-general,DBLP:conf/issac/EmirisC93,DBLP:conf/iccv/Heikkila17} suffer from the drawback that they require the input system to have as many polynomials as unknowns to be able to compute a 
resultant.
Additionally, the algorithm~\cite{DBLP:conf/iccv/Heikkila17} suffers from other drawbacks and can not be directly applied to most of the minimal problems. These drawbacks can be overcome, as we describe in the supplementary material. However, even with our proposed improvements the resultant-based method~\cite{DBLP:conf/iccv/Heikkila17}, which is based on hiding one of the input variables in the coefficient field, would result in a GEP with unwanted eigenvalues and in turn unwanted solutions to original system~\eqref{eq:system}. This leads to slower solvers for most of the studied minimal problems.

Therefore, we investigate an alternate approach where instead of hiding one of the input variables~\cite{emiris-general,DBLP:conf/iccv/Heikkila17} or using $u$-resultant of a general form~\eqref{eq:u-res}~\cite{emiris-general}, we introduce an extra variable $\lambda$ and a new polynomial of a special form, i.e., $x_{i} - \lambda$. The augmented polynomial system is solved by hiding $\lambda$ and reducing a constraint similar to~\eqref{eq:spresconstraint2} into a regular eigenvalue problem that leads to 
smaller solvers than~\cite{emiris-general,DBLP:conf/iccv/Heikkila17}. Next section lays the theoretical foundation of our approach and outlines the algorithm along with the 
steps for computing a sparse resultant matrix $\M{M}(\lambda)$.

\section{Sparse resultants using an extra equation}
\noindent We start with a set of $m$ polynomials from~\eqref{eq:system} in $n$ variables $x_1,\dots,x_n$ to be solved. Introducing an extra variable $\lambda$ we define $\V{x^{\prime}} = \left[ x_{1},\dots ,x_{n}, \lambda \right]$ and an extra polynomial $f_{m+1}(\V{x^{\prime}}) = x_{i} - \lambda$. Using this, we propose an algorithm inspired by~\cite{DBLP:conf/iccv/Heikkila17} and~\cite{emiris-general} to solve the following augmented polynomial system for $\V{x^{\prime}}$, 
\vspace{-0.2cm}
\begin{equation}\label{eq:inputaug}
    f_{1}(\V{x^{\prime}}) = 0,\dots ,f_{m}(\V{x^{\prime}}) = 0, f_{m+1}(\V{x^{\prime}}) = 0.
\end{equation}
Our idea it to compute its sparse resultant matrix $\M{M}(= \M{M}(\lambda))$ by hiding $\lambda$ in a way that allows us to solve~\eqref{eq:inputaug} by reducing its linearization (similar to~\eqref{eq:spresconstraint2}) to an eigenvalue problem. 
\subsection{Sparse resultant and eigenvalue problem}\label{subsec:spreseigprob}
\noindent Our algorithm computes the monomial multiples of the polynomials in~\eqref{eq:inputaug} in the form of a set $T = \lbrace T_{1},\dots,T_{m},T_{m+1} \rbrace$ where each $T_{i}$ denotes the set of monomials to be multiplied by $f_{i}(\V{x^{\prime}})$. We may order monomials in each $T_{i}$ 
to obtain a vector form, $\V{T}_{i}\! =\! \text{vec}(T_{i})$ and stack these vectors as
 $   \V{T} = \left[ \V{T}_{1},\dots ,\V{T}_{m},\V{T}_{m+1} \right].$
The set of all monomials present in the resulting extended set of polynomials $\lbrace \V{x}^{\V{\alpha_i}}f_i(\V{x^{\prime}}), \forall \V{x}^{\V{\alpha_i}}\in T_i, i=1,\dots m+1 \rbrace$ is called the monomial basis and is denoted as $B = \lbrace \V{x}^{\V{\alpha}} \mid \V{\alpha} \in \mathbb{Z}_{\geq 0}^{n} \rbrace$. The vector form of $B$ w.r.t. some monomial ordering is denoted as $\V{b}$. Then the extended set of polynomials can be written in a matrix form,
\begin{equation}\label{eq:spresmat}
    \M{M} \ \V{b} = 0,
\end{equation}
The coefficient matrix $\M{M}$ is a function of $\lambda$ as well as the coefficients of input polynomials~\eqref{eq:inputaug}. Let $\varepsilon = |B|$. Then by construction~\cite{DBLP:conf/iccv/Heikkila17} $\M{M}$ is a \textit{tall} matrix with $p \geq \varepsilon $ rows. We can remove extra rows and form an invertible square matrix which is the sparse resultant matrix mentioned in previous section. While \Hk~\cite{DBLP:conf/iccv/Heikkila17} solve a problem similar to~\eqref{eq:spresmat} as a GEP, we exploit the structure of newly added polynomial $f_{m+1}(\V{x^{\prime}})$ and propose a block partition of $\M{M}$ to reduce the matrix equation of~\eqref{eq:spresmat} to a regular eigenvalue problem.
\begin{proposition}\label{prop:eigendecomp}
Let  $f_{m+1}(\V{x^{\prime}}) = x_{i} - \lambda$, then there exists a block partitioning of  $\M{M}$ in~\eqref{eq:spresmat} as:
\begin{equation}\label{eq:blockpartition}
    \M{M} = \begin{bmatrix} \M{M}_{11} & \M{M}_{12} \\ \M{M}_{21} & \M{M}_{22} \end{bmatrix},
\end{equation}
such that~\eqref{eq:spresmat} can be converted to an eigenvalue problem of the form $\M{X} \ \V{b^{\prime}} = \lambda \V{b^{\prime}}$. 
\end{proposition}
\textit{Proof:}
In order to block partition the columns in~\eqref{eq:blockpartition} we need to partition $B$ as $B = B_{\lambda} \sqcup B_{c}$ where 
\begin{equationarray}{lcl}\label{eq:monomialbasispartition}
    B_{\lambda} = B \cap T_{m+1},& &B_{c} = B - B_{\lambda}.
\end{equationarray}
Let us order the monomials in $B$, such that $\V{b}\! =\! \text{vec}(B)\! =\!  \begin{bmatrix} \text{vec}(B_{\lambda})\; \text{vec}(B_{c}) \end{bmatrix}^{T}\! =\! \begin{bmatrix} \V{b}_{1}\; \V{b}_{2} \end{bmatrix}^{T}$. Such a partition of $\V{b}$ induces a column partition of $\M{M}$~\eqref{eq:spresmat}. We row partition $\M{M}$ such that the lower block is row-indexed by monomial multiples of $f_{m+1}\V{(x^{\prime})}$ which are linear in $\lambda$ (\ie $\V{x}^{\V{\alpha_j}} (x_{i} - \lambda), \V{x}^{\V{\alpha_j}}\in T_{m+1}$) while the upper block is indexed by monomial multiples of $f_{1}\V{(x^{\prime})},\dots,f_{m}\V{(x^{\prime})}$. Such a row and column partition of $\M{M}$ gives us a block partition as in~\eqref{eq:blockpartition}. As $\begin{bmatrix} \M{M}_{11}\! &\! \M{M_{12}} \end{bmatrix}$ contains polynomials independent of the $\lambda$ and $\begin{bmatrix} \M{M}_{21}\! &\! \M{M}_{22}\! \end{bmatrix}$ contains polynomials of the form $\V{x}^{\V{\alpha_j}} ( x_{i} - \lambda)$ we obtain
\begin{equationarray}{lcl}\label{eq:blockmatdecom}
\M{M}_{11} = \M{A}_{11}, \; \; \; \M{M_{12}} = \M{A}_{12} \nonumber \\
\M{M}_{21} = \M{A}_{21} + \lambda \M{B_{21}}, \; \; \;  
\M{M}_{22} = \M{A}_{22} + \lambda \M{B_{22}},
\end{equationarray}
where $\M{A}_{11}, \M{A}_{12}, \M{A}_{21}$ and $\M{A}_{22}$ are matrices dependent only on the coefficients of input polynomials in~\eqref{eq:inputaug}. We assume here that $\M{A}_{12}$ has \textit{full column rank}. Substituting \eqref{eq:blockmatdecom} in~\eqref{eq:blockpartition} gives 
\begin{equationarray}{lcll}
\M{M} &= \begin{bmatrix} \M{M}_{11} & \M{M_{12}} \\ \M{M}_{21} & \M{M}_{22} \end{bmatrix} =& \begin{bmatrix} \M{A}_{11} & \M{A}_{12} \\ \M{A}_{21} & \M{A}_{22} \end{bmatrix} + \lambda \begin{bmatrix} \M{0} & \M{0} \\ \M{B_{21}} & \M{B_{22}} \end{bmatrix}&
\end{equationarray}
We can order monomials so that $\V{T}_{m+1} = \V{b}_{1}$. Now chosen partition of $\M{M}$ implies that $\M{M}_{21}$ is column indexed by $\V{b}_{1}$ and row indexed by $\V{T}_{m+1}$. As $\begin{bmatrix} \M{M}_{21}\! &\! \M{M}_{22}  \end{bmatrix}$ has rows of form $\V{x}^{\V{\alpha_j}}(x_{i}\! -\! \lambda)$, $\V{x}^{\V{\alpha_j}}\! \in\! T_{m+1}\! \implies\! \V{x}^{\V{\alpha_j}}\! \in\! B_{\lambda}$. This gives us, $\M{B_{21}} = \M{-I}$, where $\M{I}$ is an identity matrix of size $|B_{\lambda}|$ and $\M{B_{22}}$ is a zero matrix of size $|B_{\lambda}| \times |B_{c}|$. This also means that $\M{A}_{21}$ is a square matrix of same size as $\M{B}_{21}$. Thus we have a decomposition as
\begin{equationarray}{rrl}\label{eq:spresmatstruct}
\M{M} & = & \M{M_{0}} + \lambda \M{M_{1}}  =  \begin{bmatrix} \M{A}_{11} & \M{A}_{12} \\ \M{A}_{21} & \M{A}_{22} \\ \end{bmatrix} + \lambda \begin{bmatrix} \M{0} & \M{0} \\ - \M{I} & \M{0} \end{bmatrix},
\end{equationarray}
where $\M{M}$ is a $p \times \varepsilon$ matrix. If $\M{M}$ is a \textit{tall} matrix, so is $\M{A}_{12}$ from which we can eliminate extra rows to obtain a square invertible matrix $\hat{\M{A}}_{12}$ while preserving the above mentioned structure, as discussed in Section~\ref{subsec:colred}. 
Let $\V{b} = \begin{bmatrix} \V{b}_{1} \!\! & \!\! \V{b}_{2} \end{bmatrix}^{T}$. Then from~\eqref{eq:spresmat} and~\eqref{eq:spresmatstruct} we have
\begin{equationarray}{lrcccl}\label{eq:spresmatstructdecomp}
 & \begin{bmatrix} \M{A}_{11} & \hat{\M{A}}_{12} \\ \M{A}_{21} & \M{A}_{22} \end{bmatrix} \begin{bmatrix} \V{b}_{1}  \\ \V{b}_{2} \end{bmatrix} + \lambda \begin{bmatrix} \M{0} & \M{0} \\ - \M{I} & \M{0} \end{bmatrix} \begin{bmatrix} \V{b}_{1} \\ \V{b}_{2} \end{bmatrix} & = & \V{0} & \nonumber \\
 \implies & \M{A}_{11} \V{b}_{1} + \hat{\M{A}}_{12} \V{b}_{2} &=& \V{0}, & \nonumber \\
 & \M{A}_{21} \V{b}_{1} + \M{A}_{22} \V{b}_{2} - \lambda \V{b}_{1} &=& \V{0}& 
\end{equationarray}

\noindent Eliminating $\V{b}_{2}$ from the above pair of equations we obtain
\begin{eqnarray}\label{eq:eigproblem}
  \overbrace{(\M{A}_{21} - \M{A}_{22} \hat{\M{A}}_{12}^{-1} \M{A}_{11})}^{\M{X}} \V{b}_{1} = \lambda \V{b}_{1}.
\end{eqnarray}
If $\M{A_{12}}$ does not have full column rank, we change the partitioning of columns of $\M{M}$ by changing the partitions, $B_{\lambda} \! =\! \lbrace \V{x}^{m}\! \in\!  T_{m+1}  \mid  x_{i} \V{x}^{m}  \in  B \rbrace$ and $B_{c}\! =\!  B\! -\! B_{\lambda}$ by exploiting the form of $f_{m+1}(\V{x}^{\prime})$. This gives us $\M{A}_{21}\! =\! \M{I}$ and $\M{A}_{22}\! =\! \M{0}$. It also results in a different $\M{A}_{12}$ which would have full column rank. Hence from \eqref{eq:spresmatstruct} we have 
\begin{equationarray}{rrl}\label{eq:spresmatstructalt}
\M{M} & = & \M{M_{0}} + \lambda \M{M_{1}}  =  \begin{bmatrix} \M{A}_{11} & \M{A}_{12} \\ \M{I} & \M{0} \\ \end{bmatrix} + \lambda \begin{bmatrix} \M{0} & \M{0} \\ \M{B}_{21} & \M{B}_{22} \end{bmatrix},
\end{equationarray}
which is substituted in~\eqref{eq:spresmat} to get $\M{A}_{11} \V{b}_{1} + \M{A}_{12} \V{b}_{2} = \V{0}$ and $\lambda(\M{B}_{21} \V{b}_{1} \! +\! \M{B}_{22} \V{b}_{2}) \! +\! \V{b}_{1} \! =\! \V{0}$. Eliminating $\V{b}_{2}$ from these equations we get an alternate eigenvalue formulation: 
\begin{eqnarray}\label{eq:eigproblemalt}
    (\M{B}_{21} - \M{B}_{22} \hat{\M{A}}_{12}^{-1} \M{A}_{11}) \V{b}_{1} = -(1/\lambda) \V{b}_{1}. \;\; \qedsymbol
\end{eqnarray} 
We note that~\eqref{eq:eigproblem} defines our proposed solver. Here we can extract solutions to $x_{1},\dots ,x_{n}$ by computing eigenvectors of $\M{X}$. If in case $\hat{\M{A}}_{12}$ is not invertible, we can use the alternate formulation~\eqref{eq:eigproblemalt} and extract solutions in a similar manner. It is worth noting that the speed of execution of the solver depends on the size of $\V{b}_{1}$(=$|B_{\lambda}|$) as well the size of $\hat{\M{A}}_{12}$ while the accuracy of the solver largely depends on the matrix to be inverted \ie $\hat{\M{A}}_{12}$. Hence, in next section we outline a generalized algorithm for computing a set of monomial multiples $T$ as well as the monomial basis $B$ that leads to matrix $\M{M}$ satisfying  Proposition~\ref{prop:eigendecomp}. 

\subsection{Computing a monomial basis}\label{subsec:compmonbasis}
\noindent Our approach is based on the algorithm explored in~\cite{DBLP:conf/iccv/Heikkila17} for computing a monomial basis $B$ for a sparse resultant.

We briefly define the basic terms related to convex polytopes used for computing a monomial basis $B$. A Newton polytope of a polynomial $\text{NP}(f)$ is defined as a convex hull of the exponent vectors of the monomials occurring in the polynomial (also known as the support of the polynomial). Hence, we have $\text{NP}(f_{i}) = \text{Conv}(A_{i})$ where $A_{i} = \lbrace \V{\alpha} | \V{\alpha} \in \mathbb{Z}_{\geq 0}^{n} \rbrace$ is the set of all integer vectors that are exponents of monomials with non-zero coefficients in $f_{i}$. A Minkowski sum of any two convex polytopes $P_1, P_2$ is defined as $P_1+P_2 = \lbrace p_1 + p_2 \ | \ \forall p_1 \in P_1, p_2 \in P_2 \rbrace$. An extensive treatment of polytopes can be found from \cite{Cox-Little-etal-05}. The algorithm by \Hk~\cite{DBLP:conf/iccv/Heikkila17} basically computes the Minkowski sum of the Newton polytopes of a subset of input polynomials, $Q = \Sigma_{i} \text{NP}(f_{i}(\V{x}))$. The set of integer points in the interior of $Q$ defined as $B = \mathbb{Z}_{n-1} \cap (Q + \delta)$, where $\delta$ is a small random displacement vector, can provide a monomial basis $B$ satisfying the constraint~\eqref{eq:spresconstraint2}. Our proposed approach computes $B$ as a prospective monomial basis in a similar way, albeit for a modified polynomial system~\eqref{eq:inputaug}. Next we describe our approach and provide a detailed algorithm for the same in the supplementary material.

Given a system of $m$($\geq n$) polynomials~\eqref{eq:system} in $n$ variables $X =\lbrace x_1,\dots,x_n \rbrace$ we introduce a new variable $\lambda$ and create $n$ augmented systems $F^{\prime} = \lbrace f_{1},\dots,f_{m}, x_{i} - \lambda \rbrace$ for each variable $x_i \in X$. Then we compute the support $A_{j}=\text{supp}(f_{j})$ and the Newton polytope $\text{NP}(f_{j}) = \text{conv}(A_{j})$ for each polynomial $f_{j} \in F^{\prime}$. The unit simplex $\text{NP}_{0} \subset \mathbb{Z}^{n}$ is also computed. For each polynomial system $F^{\prime}$, we consider each subset of polynomials $F_{\text{sub}} \subset F^{\prime}$ and compute its Minkowski sum, $Q = \text{NP}_{0} + \Sigma_{f \in F_{\text{sub}}} \text{NP}(f)$. Then for various displacement vectors $\delta$ we try to compute a candidate monomial basis $B$ as the set of integer points inside $Q+\delta$.
 
From $B$ we compute $T_{j} = \lbrace t \in \mathbb{Z}^{n} \mid t + \text{supp}(f_{j}) \subset B \rbrace, \forall f_{j} \in F^{\prime}$. Assuming $T$ to be the set of monomial multiples for input polynomials, our approach tests that $\Sigma_{j=1}^{m+1}|T_{j}| \geq |B| $, $\min\limits_{j}|T_{j}| > 0$ and $\textit{rank}(\M{M}) = |B|$. If successful, we compute the coefficient matrix $\M{M}$ indexed by $B$ and $T$ as in Section~\ref{subsec:spreseigprob} and partition $B$ into sets $B_{\lambda} = B \cap T_{m+1}$(or $B_{\lambda} \! =\! \lbrace \V{x}^{m}\! \in\!  T_{m+1}  \mid  x_{i} \V{x}^{m}  \in  B \rbrace$ if we need to use the alternate formulation~\eqref{eq:eigproblemalt}) and $B_{c} = B - B_{\lambda}$. If the submatrix of $\M{M}$ column indexed by $B_{c}$ and row indexed by $T_{1} \cup \dots \cup T_{m}$ has full column rank then we add $B$ to the list of favourable monomial bases.
 
Our algorithm then goes through all of the favorable monomial bases so computed and selects the smallest monomial basis $B$ among them along with the corresponding set of monomial multiples $T$ from which the coefficient matrix $\M{M}$ is constructed as described in Section~\ref{subsec:spreseigprob}. 
 
Next, we list the prominent features of our approach and how they seek to address the shortcomings of~\cite{emiris-general, DBLP:conf/iccv/Heikkila17}:
\vspace{-0.2cm}
\begin{enumerate}
\itemsep-0.3em
    \item We attempt to generate the smallest basis $B$ by testing adding an extra polynomial~\eqref{eq:u-res} of a special form $x_{i}-\lambda$ for each $i$ in $1,\dots,n$.
    \item We explicitly test for rank of $\M{M}$ for each candidate basis $B$ to ensure that we have a full rank solver. This addresses the issue of rank-deficient solvers in \cite{DBLP:conf/iccv/Heikkila17}.
    \item The partition of monomial basis, $B= B_{\lambda} \sqcup B_{c}$~\eqref{eq:monomialbasispartition}(or the alternate partition of $B$ as described in Proposition~\ref{prop:eigendecomp}) highlights our approach that leads to a favourable decomposition of the coefficient matrix $\M{M}$ as in~\eqref{eq:spresmatstruct}, for solving~\eqref{eq:spresmat} as an eigenvalue problem. This helps us compute much smaller and more stable solvers as compared to ones generated in~\cite{emiris-general,DBLP:conf/issac/EmirisC93,DBLP:conf/iccv/Heikkila17}.
    \item 
    The special form of extra polynomial aids us to construct $\M{M}$ that is largely smaller than the one constructed by general $u$-resultant approach in~\cite{emiris-general}.
    \item Our method can generate solvers for $m \geq n$ in~\eqref{eq:system}.
\end{enumerate}

\subsection{Removing columns from coefficient matrix}\label{subsec:colred}
\noindent The next step in our method is attempt to reduce the size of the coefficient matrix $\M{M}$ computed in the previous section. For this, we select columns of $\M{M}$ one by one in a random order to test for its removal. For each such column, we select rows (say $r_{1},\dots,r_{k}$) that contain non-zero entries in the column and also consider all columns (say $c_{1},\dots,c_{l}$) that have non-zero entries in $r_{1},\dots,r_{k}$. Then we can remove these $k$ rows and $l$ columns from $\M{M}$ only if the following conditions hold true for the resulting reduced matrix $\M{M_{\text{red}}}$. This also means that we would be removing monomials from $B$ that index $c_{1},\dots,c_{l}$ and removing monomials from $T$ that index $r_{1},\dots,r_{k}$.
\vspace{-0.2cm}
\begin{enumerate}
\itemsep-0.3em
    \item After eliminating the monomials from $T$, we require that there is at least one monomial left in each $T_{i}$.
    \item If $\M{M}$ is of size $p \times \varepsilon$, the reduced matrix $\M{M_{\text{red}}}$ would be of size $(p-k) \times (\varepsilon-l)$. Then we require $p-k \geq \varepsilon-l$ and $\text{rank}(\M{M_{\text{red}}}) = \varepsilon-l$.
    \item $\M{M_{\text{red}}}$ must be block partitioned and decomposed as in Proposition~\ref{prop:eigendecomp}.
\end{enumerate}
\vspace{-0.2cm} We repeat the above process until there are no more columns that can be removed. We note that the last condition is important as it ensures that at each stage, the reduced matrix can still be partitioned and decomposed into an eigenvalue formulation~\eqref{eq:eigproblem}. Now, reusing the notation, let's denote $\M{M}$ to be the reduced coefficient matrix and denote $B$ and $T$ to be reduced monomial basis and set of monomial multiples, respectively. 

If $\M{M}$ still has more rows than columns, we transform it into a square matrix by removing extra rows(say $q_{1},\dots,q_{j}$) and the monomials from $T$ indexing these rows. These rows are chosen in a way so that the three conditions mentioned above are still satisfied. Moreover, our proposed approach first tries to remove as many rows as possible from the lower block, indexed by $T_{m+1}$. This is to reduce $|T_{m+1}|$($=|B_{\lambda}|$) as much as possible and ensure that the matrix $\M{A}_{21}$ and hence $\M{X}$~\eqref{eq:eigproblem} for eigenvalue problem has as small size as possible. Then, if there are more rows still to be removed, the rest is randomly chosen from the upper block indexed by $\lbrace T_{1},\dots,T_{m} \rbrace$. Detailed algorithms for these two steps of matrix reduction are provided in the supplementary material. But we note that at the end of these two steps, we have the sparse resultant matrix, $\M{M}$ satisfying~\eqref{eq:spresmat} which is then reduced to the eigenvalue formulation~\eqref{eq:eigproblem}. 
\vspace{-0.1cm}

\section{Experiments}\label{sec:experiments}
\noindent 
We evaluate the performance of our method by comparing the stabilities as well as computational complexities of the solvers generated using our method with the state-of-art \gb solvers for many interesting minimal problems. The minimal problems selected for comparison represent a huge variety of relative and absolute pose problems and correspond to that studied in~\cite{DBLP:conf/cvpr/LarssonOAWKP18}. Results for additional  problems are provided in the supplementary material.

\begin{figure}[t]
\centering
\includegraphics[width=0.48\linewidth]{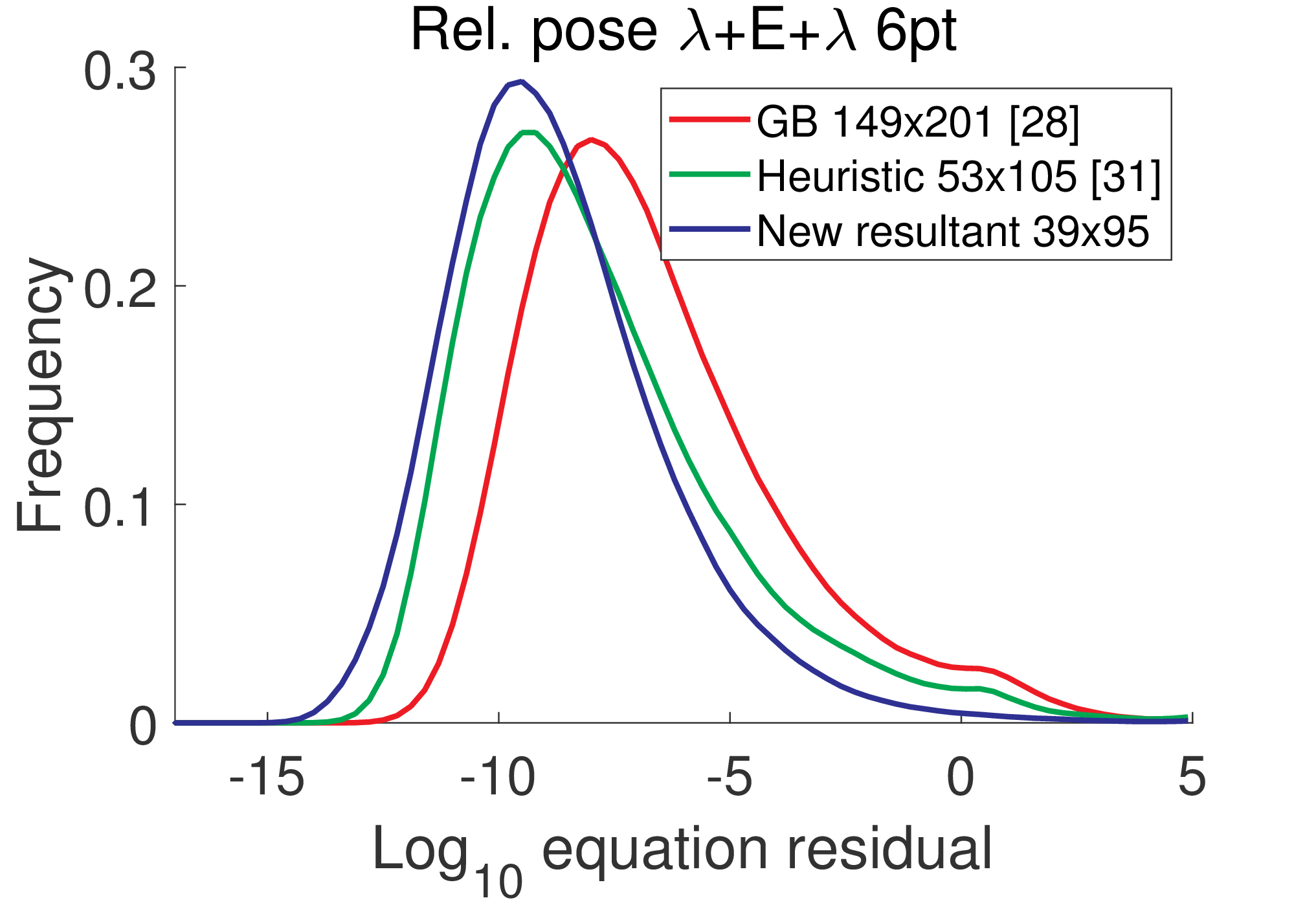}
\includegraphics[width=0.48\linewidth]{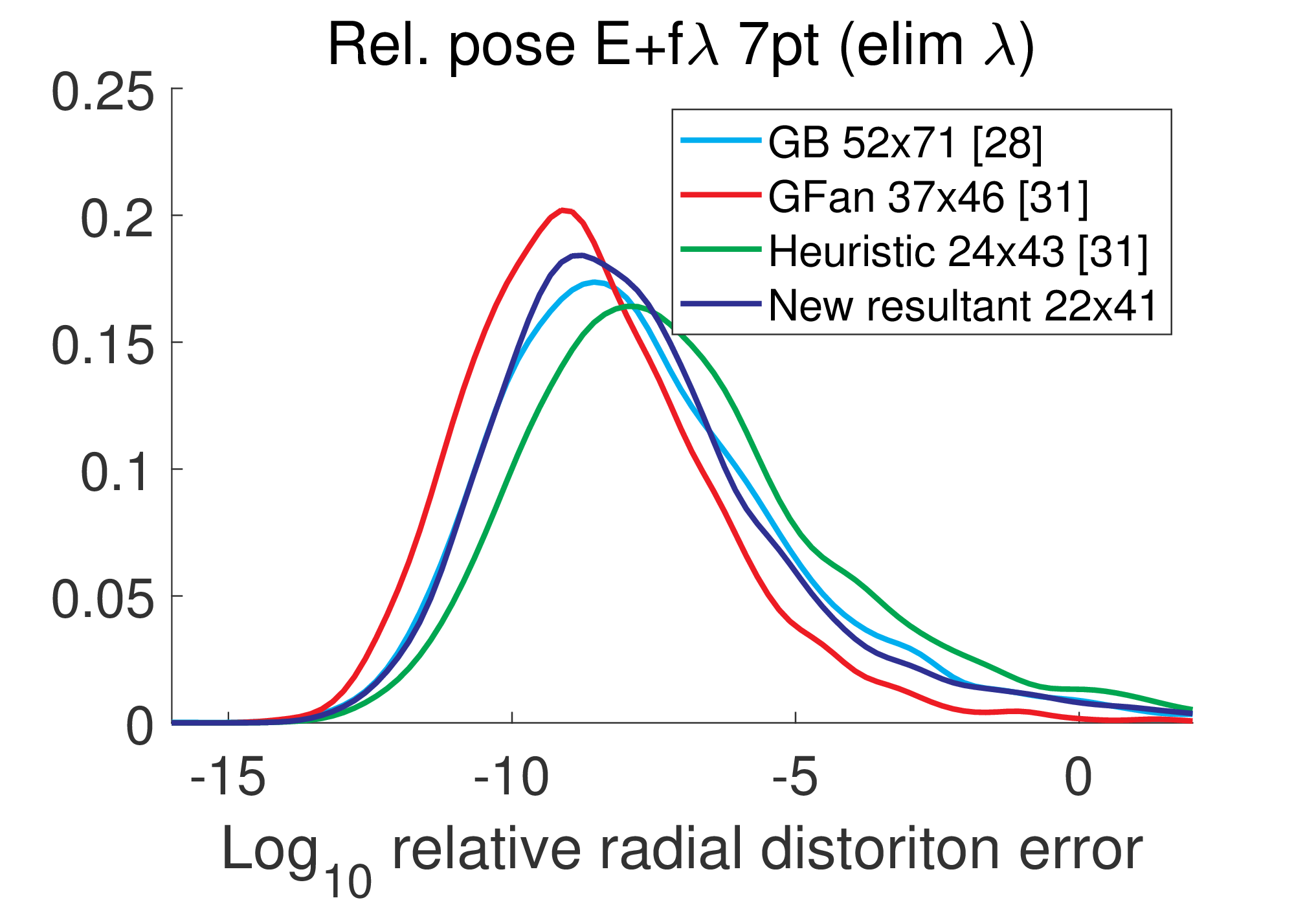}
\caption{Histograms of (left) $Log_{10}$ normalized equation residual error for Rel. pose $\lambda$ + E + $\lambda$ 6pt problem, (right) $Log_{10}$ relative error in radial distortion for Rel. pose E+$f\lambda$ 7pt (elim$\lambda$) problem. 
}
\label{fig:7pt}
\end{figure}
\vspace{-0.1cm}
\subsection{Evaluation}\label{subsec:evaluations}
\noindent The comparison of the computational complexity of minimal solvers is based on the sizes of matrix templates to be solved. E.g. a solver of size $11 \times 20$ in the table means inverting a $11 \times 11$ matrix and then a computation of $20-11=9$ eigenvalues and eigenvectors. So in Table~\ref{tbl:sizecomparison} we compare the size of templates in our resultant-based solvers with the templates used in state-of-the-art \gb solvers as well as in the original solvers proposed by the respective authors (see column $3$). The \gb  solvers used for comparison include the solvers generated using the approach in~\cite{larsson2017efficient}, the \gf and heuristic-based approaches in~\cite{DBLP:conf/cvpr/LarssonOAWKP18}. As we can see from Table~\ref{tbl:sizecomparison}, our new resultant-based approach leads to the smallest templates and hence fastest solvers for most of the minimal problems while for only a few problems our generated solver is slightly larger than the state-of-the-art solver based on the \gf or the heuristic-based method~\cite{DBLP:conf/cvpr/LarssonOAWKP18}. For some solvers though we have a slightly larger eigenvalue problem, the overall template size is considerably smaller. E.g. in the problem of estimating the relative pose and radial distortion parameter from $6$pt correspondences~\cite{Kukelova-ECCV-2008} we have an eigenvalue problem of size $56 \times 56$ and matrix inversion of size $39 \times 39$ whereas the heuristic-based solver has a $52 \times 52$ eigenvalue problem  but inversion of a larger matrix of size $53 \times 53$. For this problem the resultant-based solver is slightly faster than the state-of-the-art heuristic-based solver~\cite{DBLP:conf/cvpr/LarssonOAWKP18}. Note that for this problem we failed to generate a \gf solver~\cite{DBLP:conf/cvpr/LarssonOAWKP18} in reasonable time. 
It is worth noting that here we do not compare our solvers' sizes with resultant-based solvers generated by original versions of~\cite{DBLP:conf/iccv/Heikkila17} and~\cite{emiris-general}. These methods can not be directly applied to most of the studied minimal problems as they can not handle more equations than unknowns. With~\cite{DBLP:conf/iccv/Heikkila17} we also failed to generate full rank solvers for some problems. Even after proposing extensions to these methods~\cite{DBLP:conf/iccv/Heikkila17, emiris-general}, the generated solvers were larger than ours, and GEP involved in~\cite{DBLP:conf/iccv/Heikkila17} led also to many unwanted solutions. We give the sizes of these solvers in supplementary material along with a brief description of our proposed improvements to~\cite{DBLP:conf/iccv/Heikkila17}.

We evaluate and compare the stabilities of our solvers from Table~\ref{tbl:sizecomparison} with \gb solvers.
As it is not feasible to generate scene setups for all considered problems, we instead evaluate the stability of minimal solvers using 5K instances of random data points. Stability measures include mean and median of $Log_{10}$ of normalized equation residuals for computed solutions as well as the solvers failures as a $\%$ of 5K instances for which at least one solution has a normalized residual $>10^{-3}$. These measures on randomly generated inputs have been shown to be sufficiently good indicators of solver stabilities~\cite{larsson2017efficient}. Table~\ref{tbl:stabcomp} shows stabilities of solvers for seven minimal problems selected from Table~\ref{tbl:sizecomparison}. Figure~\ref{fig:7pt} (left) shows histogram of $Log_{10}$ of normalized equation residuals for the ``Rel.pose $\lambda$+E+$\lambda$'' problem, where our solver is not only faster, but also more stable than the state-of-the-art solvers.
The stabilities for the remaining problems as well as histograms of residuals are in the supplementary material. 
In general, our new method generates solvers that are stable with only very few failures.

\begin{figure}[t]
\centering
\includegraphics[width=0.34\linewidth]{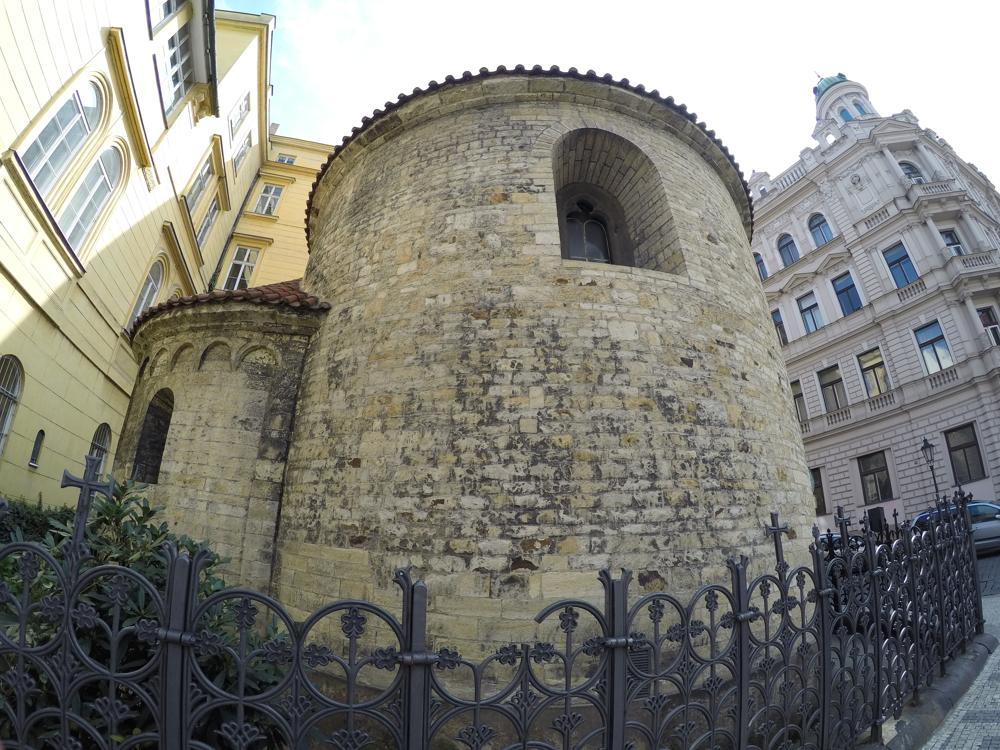}
\includegraphics[width=0.34\linewidth]{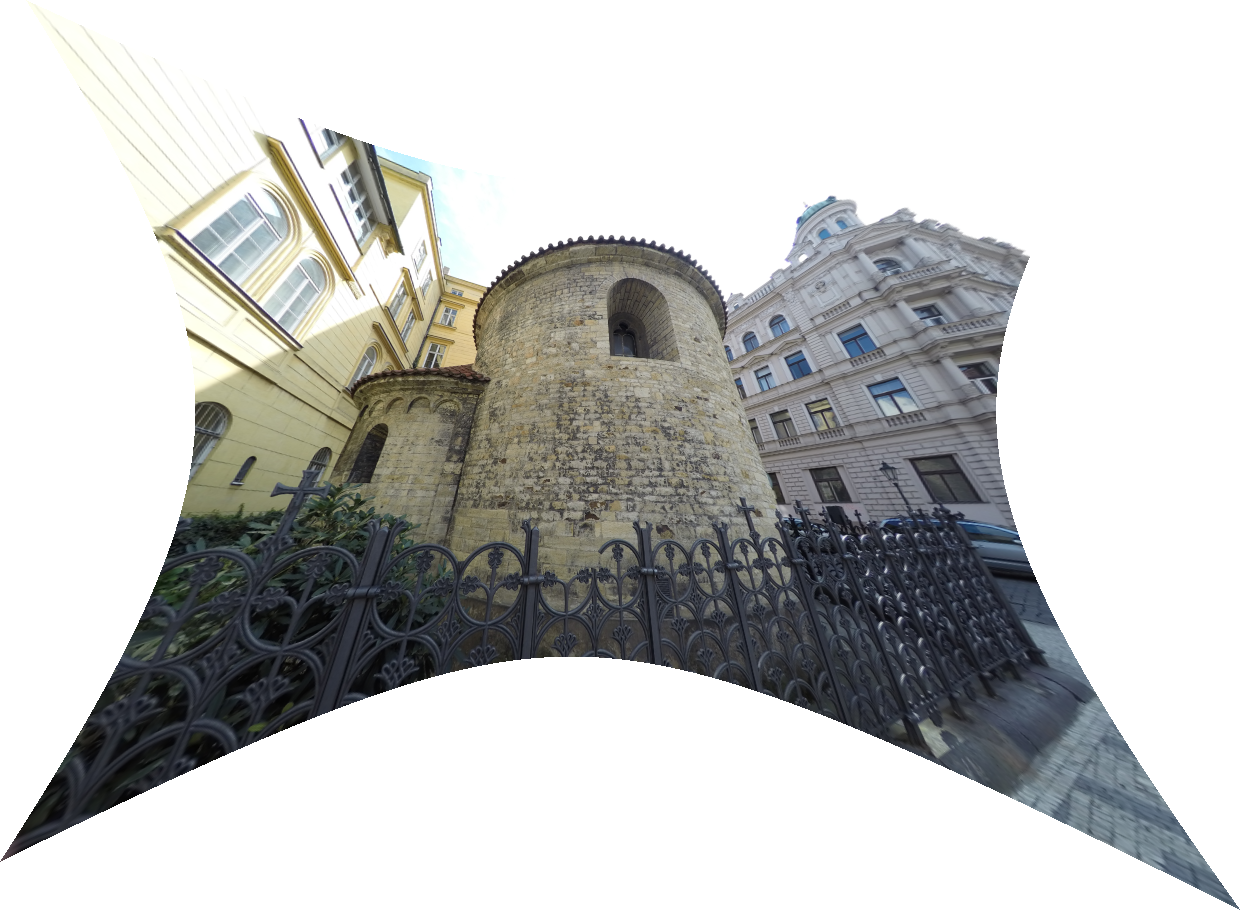}
\includegraphics[width=0.27\linewidth]{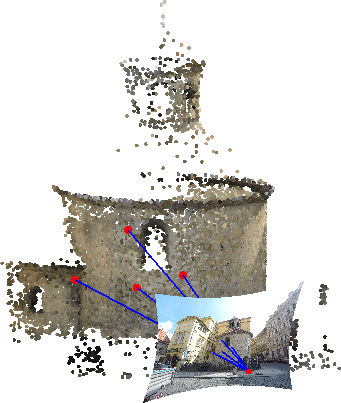} \\
\includegraphics[width=0.495\linewidth]{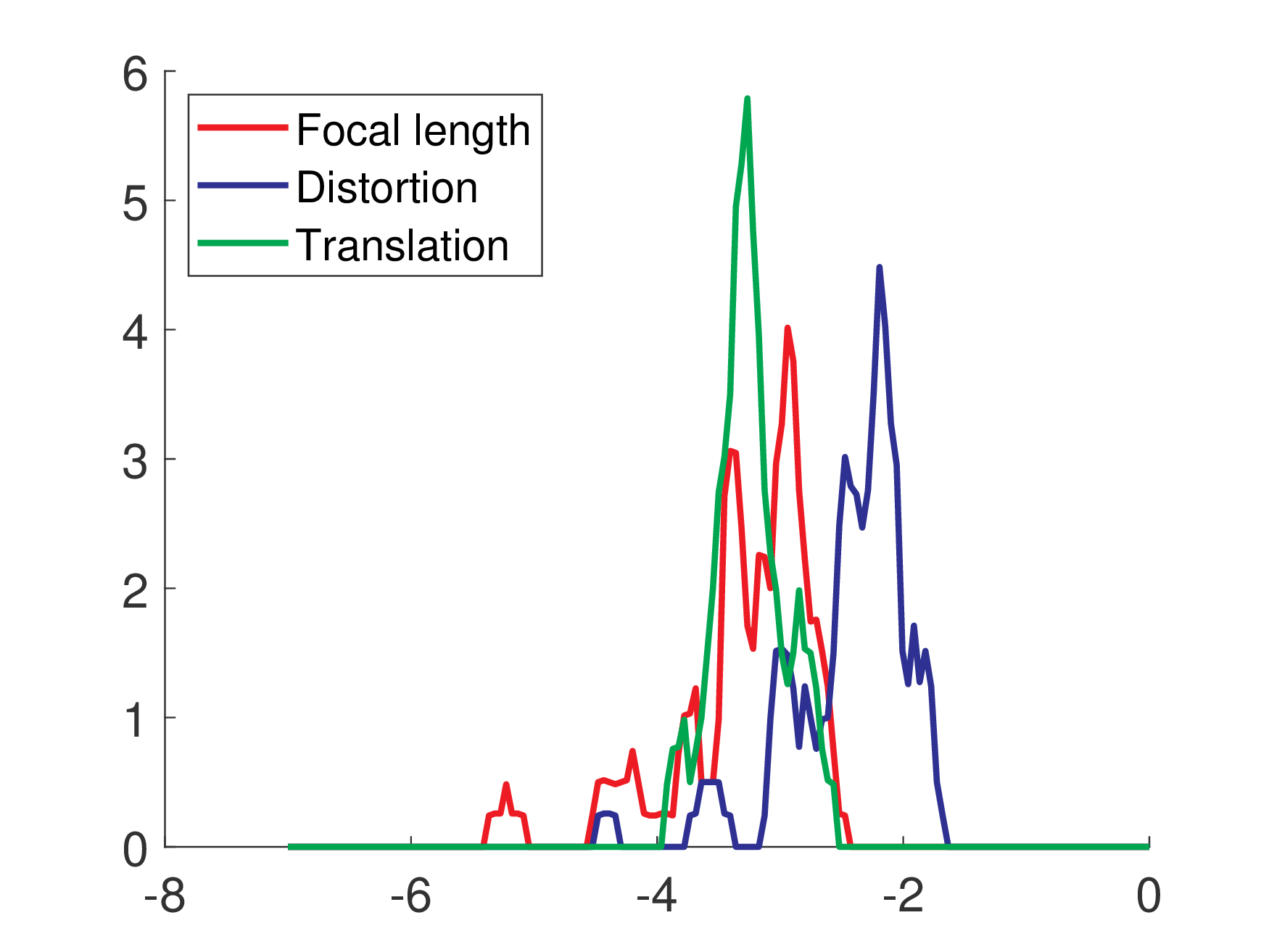}
\includegraphics[width=0.495\linewidth]{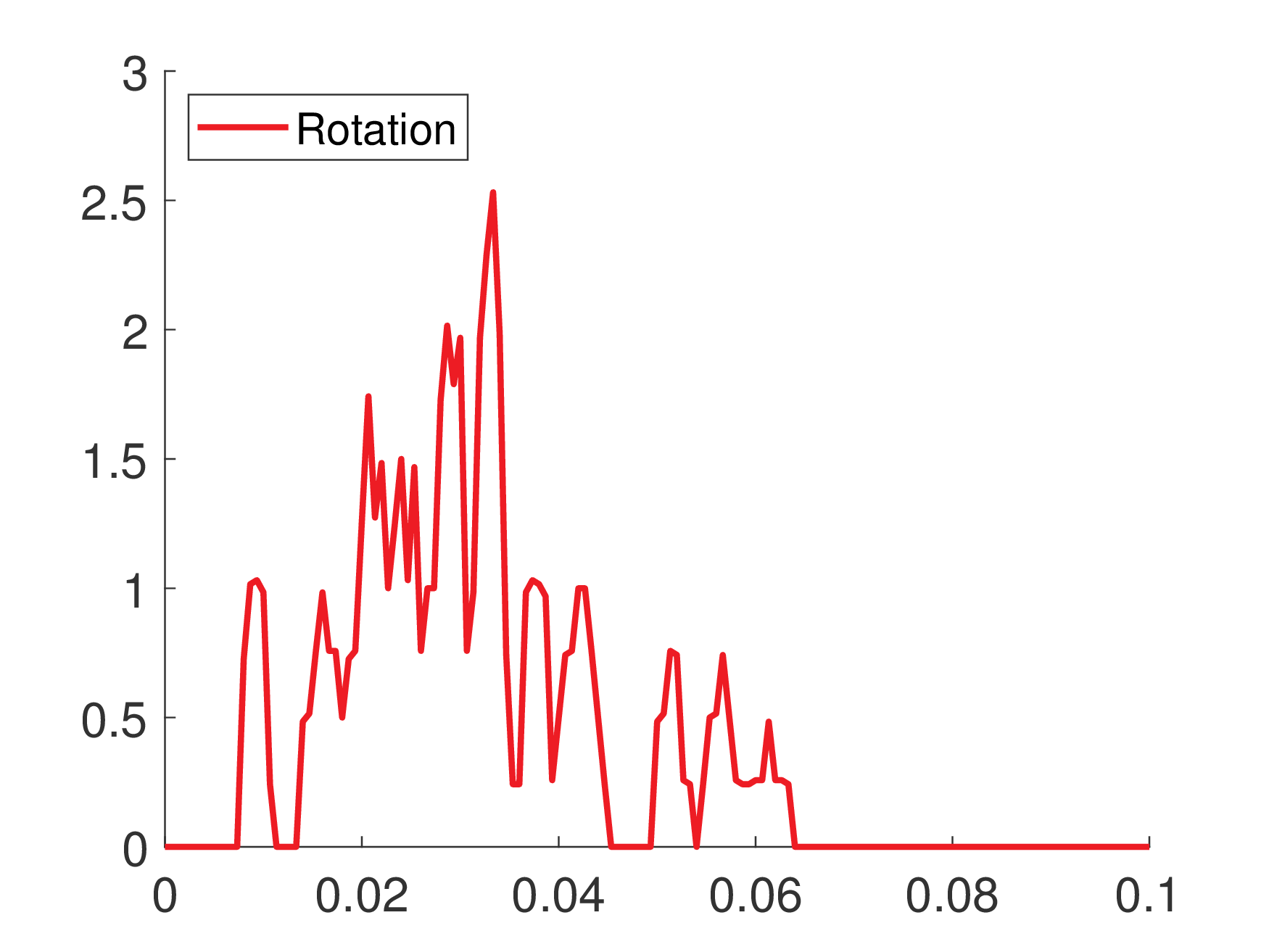}
\caption{Top row: Example of an input image (left). Undistorted image using the proposed resultant-based P4Pfr solver (middle). Input 3D point cloud and an example of registered camera (right).
Bottom row:  Histograms of errors for 62 images.
The measured errors are (left) the $Log{10}$ relative focal length $|f-f_{GT}|/f_{GT}$, radial distortion  $|k-k_{GT}|/|k_{GT}|$, and the relative translation error $\| \vec{t}-\vec{t}_{GT} \| / \|\vec{t}_{GT}\|$, and (right) the rotation error in degrees.}
\label{fig:rotunda}
\end{figure}

Note that as our new solvers are solving the same formulations of problems as the existing state-of-the-art solvers, the performance on noisy measurements and real data would be the same as the performance of the state-of-the-art solvers. The only difference in the performance comes from numerical instabilities that already appear in the noise-less case and are detailed in Table~\ref{tbl:stabcomp} (fail$\%$). For performance of the solvers in real applications we refer the reader to papers where the original formulations of the studied problems were presented (see Table~\ref{tbl:sizecomparison}, column $3$). Here we select two interesting problems, \ie one relative and one absolute pose problem, and perform experiments on synthetically generated scenes and on real images, respectively.

\begin{table*}
\centering
\begin{tabular}{l c c c c c c}
\toprule
Problem & Our & Original & \cite{larsson2017efficient}  & GFan~\cite{DBLP:conf/cvpr/LarssonOAWKP18} & (\#GB) & Heuristic~\cite{DBLP:conf/cvpr/LarssonOAWKP18} \\ \midrule
Rel. pose F+$\lambda$ 8pt\textsuperscript{$(\ddagger)$} (\small{$8$ sols.})          & $7 \times 16$ & $12\times 24$ \cite{kuang2014minimal} & $ 11\times 19 $  & $ 11\times 19 $  & $(10)$ & $\bf  7 \times 15 $  \\ 
Rel. pose E+$f$ 6pt (\small{$9$ sols.})       & $\bf 11 \times 20$ & $21 \times 30$ \cite{bujnak20093d} & $ 21\times 30 $  & $\bf  11\times 20 $  & $(66)$ & $\bf  11\times 20 $  \\ 
Rel. pose $f$+E+$f$ 6pt (\small{$15$ sols.})      & $\bf 12 \times 30$  & $31 \times 46$ \cite{Kukelova-ECCV-2008} & $ 31\times 46 $  & $ 31\times 46 $  & $(218)$ & $ 21\times 36 $  \\ 
Rel. pose E+$\lambda$ 6pt (\small{$26$ sols.})  & $\bf 14 \times 40$  & $48 \times 70 $ \cite{kuang2014minimal} & $ 34\times 60 $  & $ 34\times 60 $  & $(846)$ & $\bf  14\times 40 $  \\ 
Stitching $f\lambda$+R+$f\lambda$ 3pt (\small{$18$ sols.})  & $\bf  18\times 36 $ & $54 \times 77$ \cite{naroditsky2011optimizing}& $ 48\times 66 $  & $ 48\times 66 $  & $(26)$ & $\bf  18\times 36 $  \\ 
Abs. Pose P4Pfr (\small{$16$ sols.}) & $\bf  52\times 68 $ & $136 \times 152 $ \cite{bujnak2010new} & $ 140\times 156$  & $54\times 70 $  & $(1745)$ & $54\times 70$  \\ 
Abs. Pose P4Pfr \small{(elim. $f$)} (\small{$12$ sols.}) & $\bf  28\times 40 $ &  $\bf  28\times 40 $ \cite{larsson2017making} & $ 48 \times 60$  & $\bf 28 \times 40$  & $(699)$ & $\bf 28 \times 40$  \\ 
 Rel. pose $\lambda$+E+$\lambda$ 6pt\textsuperscript{$(\ddagger)$} (\small{$52$ sols.}) & $\bf  39\times 95$ & $238 \times 290$ \cite{Kukelova-ECCV-2008} & $ 149\times 201$  & - & ? & $  53\times 105$  \\ 
Rel. pose $\lambda_1$+F+$\lambda_2$ 9pt (\small{$24$ sols.}) & $ 90 \times 117$ & $179 \times 203$ \cite{Kukelova-ECCV-2008} & $ 189\times 213$  & $\bf  87\times 111$  & $(6896)$ & $\bf  87\times 111$  \\ 
Rel. pose E+$f\lambda$ 7pt (\small{$19$ sols.}) & $\bf  61\times 80 $   & $200\times 231 $\cite{kuang2014minimal} & $181\times 200$  & $69\times 88 $  & $(3190)$ & $  69\times 88 $  \\ 
Rel.\ pose E+$f\lambda$ 7pt \small{(elim.\ $\lambda$)} (\small{$19$ sols.}) &$\bf  22\times 41 $  & - & $ 52\times 71 $  & $ 37\times 56 $  & $(332)$ & $24\times 43 $  \\ 
Rel.\ pose E+$f\lambda$ 7pt \small{(elim.\ $f\lambda$)}  (\small{$19$ sols.}) & $\bf 51 \times 70$  & $\bf 51 \times 70$ \cite{kukelova2017clever} & $\bf  51\times 70 $  & $\bf  51\times 70 $  & $(3416)$ & $\bf  51\times 70 $  \\ 
Abs. pose quivers\textsuperscript{$(\dagger)$} (\small{$20$ sols.}) &  $ 68 \times 92$  & $372\times 386 $ \cite{kuang2013pose} & $ 203\times 223$  & - & ? & $\bf  68\times 88$  \\ 
Rel. pose E angle+4pt (\small{$20$ sols.})  & - & $ 270\times 290 $ \cite{li20134} & $ 266\times 286$  & - & ? & $\bf  183\times 203$ \\
Abs. pose refractive P5P\textsuperscript{$(\dagger)$} (\small{$16$ sols.}) & $\bf 68 \times 93$ & $280\times 399 $ \cite{haner2015absolute} & $ 199 \times 215$  & $ 112 \times 128$  & $(8659)$ & $ 199 \times 215$  \\
\bottomrule
\end{tabular}
\caption{Comparison of solver sizes for some minimal problems. Missing entries are when we failed to generate a solver. $(\dagger)$: Input polynomials were eliminated using G-J elimination before generating a solver using our resultant method as well as solvers based on~\cite{larsson2017efficient}, the \gf-based solver~\cite{DBLP:conf/cvpr/LarssonOAWKP18} and the heuristic-based solver~\cite{DBLP:conf/cvpr/LarssonOAWKP18}. $(\ddagger)$:Solved using the alternate eigenvalue formulation~\eqref{eq:eigproblemalt}.}
\label{tbl:sizecomparison}
\vspace{-2mm}
\end{table*}

\begin{table*}
\centering
\normalsize
\begin{tabular}{l c c c c c c c c c } \toprule
Problem &  \multicolumn{3} {c} {Our} & \multicolumn{3} {c} {\cite{larsson2017efficient}} & \multicolumn{3} {c} {Heuristic~\cite{DBLP:conf/cvpr/LarssonOAWKP18}}  \\ \cmidrule(r){2-4} \cmidrule(r){5-7} \cmidrule(r){8-10}
& mean & med. & fail($\%$) & mean & med. & fail($\%$) & mean & med. & fail($\%$)  \\\midrule
Rel. pose $f$+E+$f$ 6pt  &$ -12.55 $&$ -12.90 $&$\bf 0.52 $&$ -12.09$ & $-12.53$ & $ 2.36$ &$ -12.05$ &$ -12.48$ &$ 1.44$ \\
Abs. Pose P4Pfr \small{(elim. $f$)} &$-12.86$&$-13.08$&$\bf 0$ &$ -12.59$ &$ -12.85$ &$ 0 $&$ -12.73$ &$ -13.00 $&$ 0.02$    \\
Rel. pose $\lambda$+E+$\lambda$ 6pt  &$ -8.99$ &$ -9.33$ &$\bf 14.66$ &$ -6.92$ &$ -7.45$ &$ 25.9$ &$  -8.13 $&$ -8.73$ &$ 26.46$    \\
Rel. pose E+$f\lambda$ 7pt\textsuperscript{$(\ddagger)$}   &$ -11.29$ &$ -11.59$ &$\bf 0.36$   &$ -10.69$ &$ -11.13 $&$ 7.58$ & - & - & - \\ 
Rel.\ pose E+$f\lambda$ 7pt \small{(elim.\ $\lambda$)}   &$ -12.53$ &$ -12.95$ &$ 2.34$   &$ -11.99$ &$ -12.35 $&$\bf 0.44$ &$ -11.05$ &$ -11.84$ &$ 5.70$ \\
Abs. pose refractive P5P\textsuperscript{$(\dagger)$} &$ -13.03$ &$ -13.25$ &$\bf 0$ &$ -12.45$&$ -12.79$ &$ 0.10$ &$ -12.23$ &$ -12.53$ &$ 0.08$\\ \bottomrule
\end{tabular}
\caption{Stability comparison for solvers generated by our new method, solvers generated using~\cite{larsson2017efficient} and heuristic-based solvers~\cite{DBLP:conf/cvpr/LarssonOAWKP18} on some interesting minimal problems. Mean and median are computed from $Log_{10}$ of normalized equation residuals. $(\dagger)$: Solvers generated after Gauss-Jordan(G-J) elimination of input polynomials. $(\dagger)$: Failed to extract solutions to all variables for the heuristic-based solver~\cite{DBLP:conf/cvpr/LarssonOAWKP18}.}
\label{tbl:stabcomp}
\vspace{-3mm}
\end{table*}

\textbf{E+$f\lambda$ solver on synthetic scenes:} 
We study the numerical stability of the new resultant-based solver for the problem of estimating the relative pose of one calibrated and one camera with unknown focal length and radial distortion from 7-point correspondences, i.e.\ the Rel.\ pose E+$f\lambda$ 7pt problem from Table~\ref{tbl:sizecomparison}. We considered the formulation ``elim.\ $\lambda$'' proposed in~\cite{DBLP:conf/cvpr/LarssonOAWKP18} that leads to the smallest solvers. We studied the performance on noise-free data and compared it to the results of \gb solvers from Table~\ref{tbl:sizecomparison}.

We generated 10K scenes with 3D points drawn uniformly from a $\left[-10,10\right]^3$ cube. Each 3D point was projected by two cameras with random feasible orientation and position. The focal length of the first camera was randomly drawn from the interval $f_{gt} \in \left[0.5, 2.5\right]$ and the focal length of the second camera was set to $1$, i.e., the second camera was calibrated. The image points in the first camera were corrupted by radial distortion following the one-parameter division model. The radial distortion parameter $\lambda_{gt}$ was drawn at random from the interval $[-0.7, 0]$ representing distortions of cameras with a small distortion up to slightly more than GoPro-style cameras. Figure~\ref{fig:7pt} (right) shows  $Log_{10}$ of the relative error of the distortion parameter $\lambda$ obtained by selecting the real root closest to the ground truth $\lambda_{gt}$. All tested solvers provide stable results with only a small number of runs with larger errors. The new resultant-based solver (blue) is not only smaller but also slightly more stable than the heuristic-based solver from~\cite{DBLP:conf/cvpr/LarssonOAWKP18} (green). 
\textbf{P4Pfr solver on real images:} \label{sec:real_images}
\noindent We evaluated the resultant-based solver for a practical problem of estimating the absolute pose of camera with unknown focal length and radial distortion from four 2D-to-3D point correspondences, i.e. the P4Pfr solver, on real data. We consider the \textit{Rotunda} dataset, which was proposed in \cite{kukelova2015efficient} and in~\cite{larsson2017making} it was used for evaluating P4Pfr solvers. This dataset consists of 62 images captured by a GoPro Hero4 camera. 
Example of an input image from this dataset (left) as well as undistorted (middle) and registered image (right) using our new solver, is shown in Figure~\ref{fig:rotunda} (top). The Reality Capture software \cite{realitycapture} was used to build a 3D reconstructions of this scene. 
We used the 3D model to estimate the pose of each image using the new P4Pfr resultant-based solver ($28\times 40$) in a RANSAC framework. Similar to \cite{larsson2017making}, we used the camera and distortion parameters obtained from \cite{realitycapture} as ground truth for the experiment.
Figure~\ref{fig:rotunda} (bottom)
shows the errors for the focal length, radial distortion, and the camera pose. Overall the errors are quite small, \eg most of the focal lengths are within $0.1\%$ of the ground truth and almost all rotation errors are less than $0.1$ degrees, which shows that our new solver works well for real data. These results (summarized in the supplementary material) are consistent with the results of the P4Pfr solver presented in~\cite{larsson2017making}, which was tested on the same dataset. The slightly different results reported in~\cite{larsson2017making} are due to RANSAC's random nature and a slightly different P4Pfr formulation (40x50) used in~\cite{larsson2017making}.

\section{Conclusion}
\vspace{-0.2cm}
\noindent In this paper, we propose a novel algorithm for generating efficient minimal solvers based on sparse resultants, computed by adding an extra polynomial of a special form and reducing the resultant matrix constraint to an eigenvalue problem. The new approach achieves significant improvements over existing resultant-based methods in terms of efficiency of the generated
solvers.
From our experiments on many minimal problems on real and synthetic scenes,
we show that the new method is a competitive alternative to the highly optimised \gb methods.
The  fact that new resultant-based solvers have for many problems the same size as the state-of-the-are heuristic or GFan solvers, shows that these solvers are maybe already ``optimal'' w.r.t. template sizes. On the other hand, there is no one general method (GFan/heuristic/resultant), which will provably return the smallest solver for every problem and we believe that especially for complex problems all methods have to be tested when trying to generate the ``best'' solver.
\section{Acknowledgement}
The authors would like to thank Academy of Finland for the financial support of this research (grant no. 297732).
{\small
\bibliographystyle{ieee}
\bibliography{ms}
}

\end{document}


\title{Supplementary material \\
A sparse resultant based method for efficient minimal solvers
}

\author{Snehal Bhayani\\
Center for Machine Vision and Signal Analysis\\
University of Oulu, Finland\\
{\tt\small snehal.bhayani@oulu.fi}
\and
Zuzana Kukelova\\
 Center for Machine Perception\\
Czech Technical University, Prague\\
{\tt\small kukelova@cmp.felk.cvut.cz}
\and
Janne \Hk\\
Center for Machine Vision and Signal Analysis\\
University of Oulu, Finland\\
{\tt\small janne.heikkila@oulu.fi}
}

\maketitle
\section{Existing sparse resultant based algorithms}
\noindent In this section we consider the existing sparse resultant based algorithms~\cite{emiris-general, DBLP:conf/iccv/Heikkila17}, where the authors consider a system of $n$ polynomials,
\begin{eqnarray}
\label{eq:nsystem}
\lbrace f_1(x_1,...,x_n)=0,...,f_n(x_1,...,x_n)=0 \rbrace,
\end{eqnarray}
in $n$ unknowns, $X = \lbrace x_1,...,x_n\rbrace$ for computing a sparse resultant matrix. While \Hk~\cite{DBLP:conf/iccv/Heikkila17} propose a method to hide one variable, Emiris~\cite{emiris-general} propose two methods, one where they hide a variable, and another where they add an extra polynomial of the form $u_0 + u_1 x_1 + \dots + u_n x_n $, for generating a polynomial solver. In each of these methods the underlying assumption is that there are as many polynomials as there are unknowns. Hence, using their proposed algorithms we could not generate solvers for those minimal problems with more polynomials than unknowns. Additionally, the algorithm by~\cite{DBLP:conf/iccv/Heikkila17} suffers from other drawbacks as well:
\begin{enumerate} 
 \itemsep0em 
    \item Heikkil{\"{a}}~\cite{DBLP:conf/iccv/Heikkila17} propose an method of hiding one variable, say $x_{n}$, and computing a monomial basis $B$ to linearize the input polynomial equations to have
    \begin{eqnarray}\label{eq:spresconstraint}
        \M{M}(x_{n}) \V{b} = 0,
    \end{eqnarray} where $\V{b} = \text{vec}(B)$ based on some monomial order. However such a monomial basis can lead to a coefficient matrix $\M{M}(x_{n})$ that is rank deficient and hence leads to unstable or incorrect solvers.
    \item If in case $\M{M}(x_{n})$ is not rank deficient \Hk~\cite{DBLP:conf/iccv/Heikkila17} transform~\eqref{eq:spresconstraint} into a generalized eigenvalue problem(GEP) of the form
    \begin{equation}\label{eq:GEP}
        \M{A} \V{y} =  x_{n} \M{B} \V{y}.
    \end{equation} as described in (4) of Section 2.2 of our main paper. But such a conversion leads to large and sparse $\M{A}$ and $\M{B}$ that introduces parasitic eigenvalues which are either $0$ or $\infty$. It can also lead to spurious eigenvalues that correspond to incorrect solutions.
\end{enumerate}

\begin{table*}[t]
\centering
\begin{tabular}{l c c c c c}
\toprule
Problem & Extension to~\cite{DBLP:conf/iccv/Heikkila17} & \multicolumn{2} {c} {Our} & \multicolumn{2} {c} {$u$-resultant}  \\ \cmidrule(r){2-6} 
& GEP & Inv & Eig. & Inv & Eig.  \\ \midrule
Rel. pose F+$\lambda$ 8pt(\small{$8$ sols.})  & $12 \times 12$ & $\bf 11 \times 11$ & $\bf 9 \times 9$ & $15 \times 15$ & $9 \times 9$ \\
Stitching $f\lambda$+R+$f\lambda$ 3pt (\small{$18$ sols.}) & $24 \times 24$ & $\bf 18 \times 18$ & $\bf 18 \times 18$ & $31 \times 31$ & $18 \times 18$ \\  
Rel. pose E+$\lambda$ 6pt (\small{$26$ sols.})  & $30 \times 30$ & $\bf 14 \times 14$ & $\bf 26 \times 26$ & $44 \times 44$ & $26 \times 26$  \\ 
Abs. pose quivers (\small{$20$ sols.}) & $43 \times 43$ & $\bf 68 \times 68$ & $\bf 24 \times 24$  & - & -  \\ 
Rel. pose $f$+E+$f$ 6pt (\small{$15$ sols.}) & $18 \times 18$ &  $\bf 12 \times 12$ & $\bf 18 \times 18$ & - & -  \\ 
Rel. pose $\lambda_1$+F+$\lambda_2$ 9pt (\small{$24$ sols.}) & $68 \times 68$ & $\bf 90 \times 90$ & $\bf 27 \times 27$ & - & -  \\  
Rel.\ pose E+$f\lambda$ 7pt (\small{$19$ sols.}) & $36 \times 36$ & $\bf 61 \times 61$ & $\bf 19 \times 19$ & $105 \times 105 $ & $ 19 \times 19$ \\ 
Rel. pose $\lambda$+E+$\lambda$ 6pt (\small{$52$ sols.}) & $110 \times 110$ & $\bf 39 \times 39$ & $\bf 56 \times 56$ & - & -  \\ 
Triangulation from satellite im.(\small{$27$ sols.}) & $52 \times 52$ & $\bf 88 \times 88$ & $\bf  27 \times 27$ &  $ 93 \times 93$ & $27 \times 27$ \\ 
Unsynch. Rel. pose (\small{$16$ sols.}) & $128 \times 128$ & $\bf 150 \times 150$ & $\bf 18 \times 18$ & - & - \\
Rolling shutter pose (\small{$8$ sols.}) & $18 \times 18$ & $\bf 47 \times 47$ & $\bf 8 \times 8$ & $48 \times 48$ & $8 \times 8$ \\ 
\bottomrule
\end{tabular}
\caption{A comparison of the sizes of important computation steps performed by solvers generated using our new method with that of the solvers generated based on our attempted extensions of the algorithm by \Hk~\cite{DBLP:conf/iccv/Heikkila17} as well as the solvers generated using an $u$-resultant based method. Missing entry is for the case where we failed to generate a solver.}
\label{tbl:heikkilaremirissizecomp}
\end{table*}

\subsection{Proposed extension to \Hk's algorithm}
\noindent Considering the shortcomings of the method by \Hk~\cite{DBLP:conf/iccv/Heikkila17} we attempted to extend and improve their algorithm,
\begin{enumerate}
    \item Due to an iterative nature of the algorithm, it is easy to relax the requirement of having the same number of equations and unknowns, and hence we assume that there are $m \geq n$ polynomial equations with $n$ unknowns. Then we perform an exhaustive search across all polynomial combinations and variables by hiding each variable $x_i \in X$ at a time. This usually reduces the size of the monomial basis leading to a smaller matrix $\M{M}(x_{n})$ than the one generated by Heikkil{\"{a}}'s algorithm~\cite{DBLP:conf/iccv/Heikkila17}.
    
    \item The problem of rank deficiency is resolved by testing for rank of the matrix $\M{M}(x_{n})$ for every prospective monomial basis $B$ so chosen in the algorithm. This guarantees that the eigenvalues and eigenvectors of GEP formulation provides correct solutions to the original polynomial system~\eqref{eq:nsystem}.
    
    \item Additionally, we know that a GEP formulation for many minimal problems in computer vision has parasitic zero(or $\infty$) eigenvalues due to zero columns in $\M{A}$(or $\M{B}$) in~\eqref{eq:GEP}. Hence we extended the the algorithm by \Hk~\cite{DBLP:conf/iccv/Heikkila17} to eliminate a set of rows-columns in order to reduce the size of GEP we are trying to solve.
\end{enumerate}
The sizes of solvers generated using these extensions to the algorithm by \Hk~\cite{DBLP:conf/iccv/Heikkila17} for some interesting minimal problems are listed in Table~\ref{tbl:heikkilaremirissizecomp}(Column 1). If in these solvers $\M{A}$ or $\M{B}$ in GEP~\eqref{eq:GEP} is an invertible matrix, GEP can be executed as a sequence of a matrix inverse and an eigendecomposition of the resulting matrix. For example, a GEP of size $18 \times 18$ means an inverse of $18 \times 18$ matrix and an eigenvalue decomposition of $18 \times 18$ matrix. We note that this assumption holds true for all of the minimal problems in Table~\ref{tbl:heikkilaremirissizecomp}. 
In such a case the most computationally expensive step is the eigenvalue decomposition, since the matrix that is inverted is usually sparse. Now, it can be seen that for most of the minimal problems our proposed solvers are solving substantially smaller eigenvalue problems than the solvers based on the extended version of~\cite{DBLP:conf/iccv/Heikkila17}. And even though for few minimal problems the matrices to invert in our proposed solvers are slightly larger than the inverses in solvers based on~\cite{DBLP:conf/iccv/Heikkila17}, these matrices are usually quite sparse and the size difference is not as dominating as the difference in size of eigenvalue problem. 
Additionally, a GEP would lead to parasitic eigenvalues corresponding to incorrect solutions and extra computation has to be carried out in order to eliminate such eigenvalues, thus slowing down such solvers even further as compared to the ones based on our method. 
Additionally the number of eigenvalues to be computed for a GEP still is quite large as compared to the eigenvalues to be computed by our proposed solver. Hence based on these considerations, we can conclude that our proposed solvers for all of the problems in Table~\ref{tbl:heikkilaremirissizecomp} would be faster than the ones generated using our proposed extensions to~\cite{DBLP:conf/iccv/Heikkila17}.

\subsection{Comparison with Emiris's $u$-resultant method}
\noindent 
Now we consider the $u$-resultant based method~\cite{emiris-general} where the authors add a polynomial of a general form $u_0 + x_1 u_1 + \dots + x_n u_n$ with random coefficients, to the original equation~\eqref{eq:system}. However we note that in general the method presented in~\cite{emiris-general} does not work for a system with more polynomial equations than unknowns. Moreover, there is no publicly available code for the method~\cite{emiris-general}. Therefore, for a fair comparison with our method that based upon adding a polynomial of a special form, we modified our resultant-based method to simulate the one from~\cite{emiris-general}.
For this, we augmented~\eqref{eq:system} with a polynomial of the form $u_0 + x_1 u_1 + \dots + x_n u_n$ by selecting $u_1,\dots,u_n$ randomly from $\mathbb{Z}$ (for more details on $u$-resultant we refer to~\cite{emiris-general, cox2006using}). The column 3 in Table~\ref{tbl:heikkilaremirissizecomp} lists the sizes of solvers generated in this manner and is compared with the sizes of solvers generated based on our proposed method. We can observe that for many minimal problems the size of matrix to be inverted based on general $u$-resultant method is larger than that of the matrix to be inverted in our proposed solver. This indicates that our proposed solver would be faster than the solvers based on general $u$-resultant method for such minimal problems. Beyond this, for several minimal problems (5 problems from Table~\ref{tbl:heikkilaremirissizecomp}), we either failed to generate a working solver by using the above mentioned general $u$-polynomial at all or within a reasonable amount time by testing polynomial combinations of a reasonable size. We refer to Algorithm~\ref{alg:favourablebasisextraction} here and Section 3.2 of our main paper for more details about the iterative nature adopted for testing polynomial combinations of various sizes. 

Additionally we also considered the problem from computational biology explored in~\cite{emiris-general}. We compare the size of the  $u$-resultant based solver for this problem reported in~\cite{emiris-general}, with the size of a solver generated using our proposed method. This problem consists of $3$ polynomial equations in $3$ variables with $15$ generic coefficients. For more details of the algebraic problem formulation, we refer to Section 7 in~\cite{emiris-general}. Now, the mixed volume of the input polynomial system is $16$ which denotes the actual number of solutions to this polynomial system. The solver considered in~\cite{emiris-general} is generated using the $u$-resultant method by adding an extra polynomial of the form, $f_{0} = u + 31 x_{1} - 41 x_{2} + 61 x_{3}$. The solver consists of an inverse of matrix of size $56 \times 56$ and an eigenvalue decomposition of $30 \times 30$ matrix. 
We generated a solver for the same algebraic formulation with our proposed algorithm. Our new solver includes a matrix inversion of smaller matix of size $48 \times 48$ as well as smaller eigenvalue problem of size $16 \times 16$. This shows that the solver generated using our proposed algorithm would be faster than the one considered in~\cite{emiris-general}.

\section{Algorithms}
\noindent Now we consider the main contribution of our main paper for which we described a three step procedure that leads to an eigenvalue formulation(Equations (14) or (16) in our main paper) to be solved for extracting roots to~\eqref{eq:system}. So here we provide algorithms for each of these three steps. For the sake of this section, we assume details and notations of Section 3 of our main paper. We also consider a set of monomial multiples $T$ to be of form $\lbrace T_{1},\dots,T_{m} \rbrace$ where each $T_{i}$ represents the set of monomial multiples for polynomial $f_{i}(x_{1},\dots,x_{n})$. Additionally, we shall assume that wherever required a coefficient matrix $\M{M}$ is computed from a basis $B$ along with a corresponding set of monomial multiples $T$, following the lines of Section 3. With these details in mind, we now outline Algorithm~\ref{alg:favourablebasisextraction} for computing a monomial $B$ basis from a set of $m$ polynomial equations,
\begin{eqnarray}\label{eq:system}
\lbrace f_1(x_1,\dots,x_n)=0,\dots,f_m(x_1,\dots,x_n)=0 \rbrace
\end{eqnarray}
in $n$ variables. The output of the algorithm also contains a set of monomial multiples, $T$ as well as the coefficient matrix computed from $B$ and $T$. For details about the underlying theory, we refer to Section 3.1 in our main paper.
\begin{algorithm}[hbt]
    \caption{Extracting favourable monomial basis using extra equation}
    \label{alg:favourablebasisextraction}
    \hspace*{\algorithmicindent} \textbf{Input} $F = \lbrace f_{1}(\V{x}),\dots ,f_{m}(\V{x}) \rbrace$, $\V{x} = \left[ x_{1},\dots ,x_{n} \right]$ \\
    \hspace*{\algorithmicindent} \textbf{Output} $B, T, \M{M}$
    \begin{algorithmic}[1]
    \STATE{$B \gets \phi, T \gets \phi$}
    \FOR{$i \in \lbrace 1,\dots ,n \rbrace$}
    \STATE{$ F^{\prime} \gets \lbrace f_{1},\dots ,f_{m+1} \rbrace$, $f_{m+1} = x_{i} - \lambda$}
    \STATE{Calculate the support of the input polynomials: \\ $A_{j} \gets \text{supp}(f_{j}), j = 1,\dots ,m+1$}
    \STATE{Construct newton polytopes: \\ $NP_j \gets \text{conv}(A_{j}), j = 1, \dots, m+1$ as well as a unit simplex $NP_0 \subset \mathbb{Z}^{n}$.}
    \STATE{Enumerate combinations of indices of all possible sizes: \\ $K \gets \lbrace \lbrace k_{0},\dots ,k_{i} \rbrace \mid\! \forall 0 \! \leq\! i \leq (m+1); k_{0},\dots ,k_{i} \in \lbrace 0,\dots,m+1 \rbrace; k_{j} < k_{j+1}  \rbrace$}
    \STATE{Let $\Delta \gets \lbrace \lbrace \delta_{1},\dots ,\delta_{n+1} \rbrace \mid \delta_{i} \in \lbrace -\epsilon, 0, \epsilon \rbrace; i = 1,\dots ,(n+1) \rbrace$ denote the set of possible displacement vectors}
    \FOR{$I \in K$}
        \STATE{Compute the minkowski sum, $Q \gets \sum_{j \in I} (NP_{j})$}
        \FOR{$\delta \in \Delta$}
            \STATE{$B^{\prime} \gets \mathbb{Z}^{n} \cap (Q + \delta)$}
            \STATE{$T^{\prime}_{j} \gets \!  \lbrace t \in \mathbb{Z}^{n} \mid t + A_{j} \subset B^{\prime} \rbrace, j\! =\! 1\dots m+1$}
            \STATE{$T^{\prime} \gets \! \lbrace T^{\prime}_{1} \dots T^{\prime}_{m+1} \rbrace$}
            \STATE{$B^{\prime}_{\lambda} \gets \! B^{\prime} \cap T^{\prime}_{m+1}$, $B^{\prime}_{c} \gets B^{\prime} - B^{\prime}_{\lambda}$}
            \STATE{Compute $\M{M}^{\prime}$ from $B^{\prime}$ and $T^{\prime}$}
            \IF{$\Sigma_{j=1}^{m+1}|T^{\prime}_{j}| \! \geq \! |B^{\prime}|$ and $\min\limits_{j}|T^{\prime}_{j}| \! > \! 0$ and $\textit{rank}(\M{M}^{\prime}) \! = \! |B^{\prime}|$}
                \STATE{$\M{A_{12}} \gets$ submatrix of $\M{M}^{\prime}$ column indexed by $B^{\prime}_{c}$ and row indexed by $T^{\prime}_{1} \cup \dots \cup T^{\prime}_{m}$}
                \IF{$\text{rank}(\M{A_{12}}) = |B^{\prime}_{c}|$ and $ | B | \geq |B^{\prime}|$}
                        \STATE{$B \gets B^{\prime}, T \gets T^{\prime}$}
                \ENDIF
            \ENDIF
        \ENDFOR
    \ENDFOR
    \ENDFOR
    \STATE{Compute $\M{M}$ from $B$ and $T$}
    \end{algorithmic}
\end{algorithm}
For an alternate eigenvalue formulation(Equation (16) in our main paper), we need to change Step 14 in Algorithm~\ref{alg:favourablebasisextraction} to $B^{\prime}_{\lambda} \gets \lbrace \V{x}^{m} \in T^{\prime}_{m+1} \mid x_{i}\V{x}^{m} \in B^{\prime} \rbrace$, $B^{\prime}_{c} \gets B^{\prime} - B^{\prime}_{\lambda}$. 
\subsection{Removing columns from $\M{M}$}
\noindent The next step in our proposed method is to reduce the monomial basis $B$ by removing columns from $\M{M}$ along with a corresponding set of rows. A brief procedure for this step is described in Section 3.3 of our main paper, while the Algorithm~\ref{alg:basisred}, listed here achieves this. The input is the monomial basis $B$ and the set of monomial multiples $T$ computed by Algorithm~\ref{alg:favourablebasisextraction} and the output is a reduced monomial basis $B_{\text{red}}$ and a reduced set of monomial multiples, $T_{\text{red}}$ that index the columns and rows of the reduced matrix $\M{M}_{\text{red}}$ respectively. We note that this algorithm is the same irrespective of the version of eigenvalue formulation to be considered(Equations (14) or (15) in our main paper).
\begin{algorithm}[hbt]
    \caption{Reducing the monomial basis}
    \label{alg:basisred}
    \hspace*{\algorithmicindent} \textbf{Input:} $B, T$ \\
    \hspace*{\algorithmicindent} \textbf{Output:} $B_{\text{red}}, T_{\text{red}}, \M{M}_{\text{red}}$
    \begin{algorithmic}[1]
        \STATE{$B^{\prime} \gets B, T^{\prime} \gets T$}
        \REPEAT
        \STATE{$\text{stopflag} \gets$ True}
        \STATE{Compute $\M{M}^{\prime}$ from $B^{\prime}$ and $T^{\prime}$}
        \FOR{column $c$ in $\M{M}^{\prime}$}
            \STATE{Copy $\M{M}^{\prime}$ to $\M{M}^{\prime \prime}$}
            \STATE{Remove rows $r_{1},\dots,r_{k}$ containing $c$ from $\M{M}^{\prime \prime}$}
            \STATE{Remove columns $c_{1},\dots,c_{l}$ of $\M{M}^{\prime \prime}$ present in $r_{1},\dots,r_{k}$}
            \IF{$\M{M}^{\prime \prime}$ satisfies Proposition 3.1}
                \STATE{Remove monomials from $B^{\prime}$ indexing columns $c_{1},\dots,c_{l}$ }
                \STATE{Remove monomials from $T^{\prime}$ indexing rows $r_{1},\dots,r_{k}$}
                \STATE{$\text{stopflag} \gets$ False}
                \STATE{\textbf{break}}
            \ENDIF
        \ENDFOR
        \UNTIL{$\text{stopflag}$ is True}
        \STATE{$B_{\text{red}} \gets B^{\prime}, T_{\text{red}} \gets T^{\prime}$}
        \STATE{Compute $\M{M}_{\text{red}}$ from $B_{\text{red}}$ and $T_{\text{red}}$}
    \end{algorithmic}
    \end{algorithm} 
    
    Now, it may happen that the reduced matrix $\M{M}_{\text{red}}$ still has more rows than columns. Hence in our main paper, we have outlined an idea to remove excess rows so as to transform $\M{M}_{\text{red}}$ into a square matrix to facilitate a decomposition of resultant matrix constraint to an eigenvalue formulation of equation (14)(or the alternate eigenvalue formulation of equation (16). For more details we refer to Proposition 3.1 in our main paper). Towards this we provide Algorithm~\ref{alg:excessrowremoval} to remove the extra rows from $\M{M}_{\text{red}}$ by removing some monomial multiples from $T_{\text{red}}$. It accepts $B_{\text{red}}$ and $T_{\text{red}}$ as input and returns a set of monomial multiples, $T_{\text{sq}}$ that along with the basis $B_{\text{red}}$, leads to square matrix $\M{M}_{\text{sq}}$. For an alternate eigenvalue formulation(Equation (16) in our main paper), we just need to change Step 16 in Algorithm~\ref{alg:excessrowremoval} to $B^{\prime}_{\lambda} \gets \lbrace \V{x}^{m} \in T^{\prime}_{m+1} \mid x_{i}\V{x}^{m} \in B^{\prime} \rbrace$, $B^{\prime}_{c} \gets B^{\prime} - B^{\prime}_{\lambda}$. 
 \begin{algorithm}[hbpt]
    \caption{Removal of excess rows}
    \label{alg:excessrowremoval}
    \hspace*{\algorithmicindent} \textbf{Input} $B_{\text{red}}, T_{\text{red}}$\\
    \hspace*{\algorithmicindent} \textbf{Output} $T_{\text{sq}}, \M{M}_{\text{sq}}$
    \begin{algorithmic}[1]
    \STATE{$T_{\text{red}}$ contains $\lbrace T^{\prime}_{1},\dots ,T^{\prime}_{m+1} \rbrace$}
    \STATE{$B_{N} \gets |B_{\text{red}}|, T_{N} \gets \Sigma_{j=1}^{m+1}|T^{\prime}_{j}|, t_{\text{chk}} \gets \phi$}
    \WHILE{$T_{N} > B_{N}$}
        \STATE{$B^{\prime} \gets B_{\text{red}}, T^{\prime} \gets T_{\text{red}}$}
        \STATE{$T^{\prime} $ contains $\lbrace T^{\prime}_{1},\dots ,T^{\prime}_{m+1} \rbrace$}
        \STATE{Randomly select $t \in\! \lbrace t_{m} \in T^{\prime}_{m+1}\! \mid\! (t_m,m+1) \notin t_{\text{chk}} \rbrace$}
        \IF{$t$}
            \STATE{$T_{m+1}^{\prime} \gets T^{\prime}_{m+1} - \lbrace t \rbrace, T^{\prime} \gets  \lbrace T^{\prime}_{1},\dots, T_{m+1}^{\prime} \rbrace$}
            \STATE{$t_{\text{chk}} \gets t_{\text{chk}} \cup \lbrace (t,m+1) \rbrace$}
        \ELSE
            \STATE{Randomly select $i \in \lbrace 1,\dots ,m \rbrace$}
            \STATE{Randomly select $t \in  \lbrace t_{i} \in T^{\prime}_{i} \mid (t_i,i) \notin t_{\text{chk}} \rbrace$}
            \STATE{$T_{i}^{\prime} \gets T^{\prime}_{i} - \lbrace t \rbrace, T^{\prime} \gets  \lbrace T^{\prime}_{1},\dots, T^{\prime}_{m+1} \rbrace$}
            \STATE{$t_{\text{chk}} \gets t_{\text{chk}} \cup \lbrace (t,i) \rbrace$}
        \ENDIF
        \STATE{$B_{\lambda}^{\prime} \gets B^{\prime} \cap T^{\prime}_{m+1}$, $B_{c}^{\prime} \gets B^{\prime} - B_{\lambda}^{\prime}$}
        \STATE{Compute $\M{M}^{\prime}$ from $B^{\prime}$ and $T^{\prime}$}
        \IF{$\min\limits_{j}|T^{\prime}_{j}| > 0$ and $\textit{rank}(\M{M}^{\prime}) = |B^{\prime}|$}
        \STATE{$\M{A_{12}} \gets$ submatrix of $\M{M}^{\prime}$ column indexed by $B_{c}^{\prime}$ and row indexed by $T^{\prime}_{1} \cup \dots \cup T^{\prime}_{m}$}
        \IF{$\text{rank}(\M{A_{12}}) = |B_{c}^{\prime}|$}
            \STATE{$T_{\text{red}} \gets T^{\prime}, T_{N} \gets \Sigma^{m+1}_{j= 1} |T^{\prime}_{j}|$}
        \ENDIF
    \ENDIF
\ENDWHILE
\STATE{$T_{\text{sq}} \gets T_{\text{red}}$}
\STATE{Compute $\M{M}_{\text{sq}}$ from $B_{\text{red}}$ and $T_{\text{sq}}$}
\end{algorithmic}
\end{algorithm}

\section{Experiments}
\noindent In Table~\ref{tbl:sizecomparison} we provide a comparison of solvers' sizes for some additional interesting minimal problems. We can see from the table, that for all considered minimal problems  our proposed method generates the smallest solvers (sometimes of the same size as \gb solvers generated with methods from~\cite{larsson2017efficient,DBLP:conf/cvpr/LarssonOAWKP18}). For an interpretation of the solver sizes, we refer to Section 4.1 of Evaluation in our main paper. We also note that, for two of the problems in Table~\ref{tbl:sizecomparison}, we failed to generate a solver using the \gf method~\cite{DBLP:conf/cvpr/LarssonOAWKP18} in a reasonable amount of time. 
\begin{table*}
\centering
\scalebox{0.95}{
\begin{tabular}{l c c c c c c}
\toprule
Problem & Our & Original & \cite{larsson2017efficient}  & GFan~\cite{DBLP:conf/cvpr/LarssonOAWKP18} & (\#GB) & Heuristic~\cite{DBLP:conf/cvpr/LarssonOAWKP18} \\ \midrule
Rolling shutter pose (\small{$8$ sols.}) & $\bf 47 \times 55$ & $48\times 56$ \cite{saurer2015minimal} & $\bf  47 \times 55$   & $\bf  47 \times 55$   & ($520$) & $\bf  47 \times 55 $  \\ 
Triangulation from satellite im. (\small{$27$ sols.}) & $\bf 87 \times 114$ & $93\times 120$ \cite{Zheng2015satimag} & $88 \times 115$   & $88 \times 115$   & ($837$) & $88 \times 115$  \\ 
Optimal pose 2pt v2 (\small{$24$ sols.}) & $\bf 176 \times 200$ & $192 \times 216$\cite{svarm2017city} & $192 \times 216$ & $-$ & ? & $192 \times 216$  \\ 
Optimal PnP (Cayley) (\small{$40$ sols.}) & $\bf 118 \times 158$ & $\bf 118 \times 158$ \cite{DBLP:conf/bmvc/Nakano15} & $\bf 118 \times 158$   & $\bf 118 \times 158$   & $(2244)$ & $\bf 118 \times 158$  \\ 
Optimal PnP (Hesch) (\small{$27$ sols.}) & $\bf 87 \times 114$ & $93\times 120$ \cite{Hesch2011} & $88 \times 115$   & $88 \times 115$   & ($837$) & $88 \times 115$  \\ 
Unsynch. Rel. pose (\small{$16$ sols.}) & $\bf 150 \times 168 $& $633 \times 649$\cite{DBLP:journals/corr/AlblKFHSP17} & $467\times 483$ & - & ? & $299 \times 315$  \\ 
\bottomrule
\end{tabular}}
\caption{Comparison of sizes of solvers for some more minimal problems. Missing entries are when we failed to generate a \gf solver in reasonable time.}
\label{tbl:sizecomparison}
\end{table*}

Table~\ref{tbl:stabcomp} performs a stability comparison of the solvers for minimal problems from Table~\ref{tbl:sizecomparison} as well as for the problems from our main paper that were considered for comparison of sizes but were left out from the stability comparison due to the lack of space in the main paper. Just as in our main paper we measure the mean and median of $Log_{10}$ of the normalized equation residuals for computed solutions as well as the solvers failures as a \% of 5K instances for which at least one solution
has a normalized residual $> 10^{-3}$. Then our observation from the stability comparisons in Table 2 of the main paper is corroborated with our observations here for these extra set of minimal problems in Table~\ref{tbl:stabcomp}. We notice that here as well, most of the solvers based on our proposed method are similarly or more stable than the ones based on \gb methods~\cite{larsson2017efficient, DBLP:conf/cvpr/LarssonOAWKP18} and with less failures.
\begin{table*}
\centering
\normalsize
\begin{tabular}{l c c c c c c c c c } \toprule
Problem &  \multicolumn{3} {c} {Our} & \multicolumn{3} {c} {\cite{larsson2017efficient}} & \multicolumn{3} {c} {Heuristic~\cite{DBLP:conf/cvpr/LarssonOAWKP18}}  \\ \cmidrule(r){2-4} \cmidrule(r){5-7} \cmidrule(r){8-10}
& mean & med. & fail($\%$) & mean & med. & fail($\%$) & mean & med. & fail($\%$)  \\\midrule
Rel. pose F+$\lambda$ 8pt  & $-14.26$ & $-14.43$ & $\bf 0$ & $-13.74$ & $-14.26$ & $0.14$ &  $-14.18$ & $-14.48$ & $\bf 0$   \\
Rel. pose E+$f$ 6pt  & $ -13.17$ & $-13.44$ & $\bf 0$ & $-12.87$ & $-13.17$ & $0$ & $-13.05$ & $-13.34$ & $\bf 0$ \\
Rel. pose E+$\lambda$ 6pt  &$ -11.65$ &$ -11.94$ &$\bf 0.34$  &$ -11.42$ &$ -11.72$ &$ 0.52$ &$ -11.34$  &$ -11.68$  &$ 0.94$   \\
Stitching $f\lambda$+R+$f\lambda$ 3pt  &$ -13.22$ &$ -13.42$ &$\bf 0 $&$ -13.06$ &$ -13.37$ &$  0.16$ &$ -13.20$ &$ -13.46$ &$ 0.02$    \\
Rel. pose $\lambda_1$+F+$\lambda_2$ 9pt  &$ -9.81$ &$ -10.08$ &$\bf 3.32 $&$  -9.81$ &$ -10.39$ &$ 5.14$ &$  -9.56$ &$ -9.98$ &$ 6.10$ \\
Rel.\ pose E+$f\lambda$ 7pt \small{(elim.\ $f\lambda$)}  &$ -10.71$ &$ -10.95$ &$ 0.38$   &$ -10.57$ &$ -10.90 $&$\bf 0.30$ &$ -11.04$ &$ -11.32 $&$ 0.32$  \\
Abs. pose quivers\textsuperscript{$(\dagger)$}  &$  -12.39$&$ -12.60$ &$\bf 0 $ &$ -11.18$ &$ -11.51$ &$ 0.32$ &$ -12.48$ &$ -12.88$ &$\bf 0$ \\
Rolling shutter pose &$  -12.16$&$ -12.34$ &$ \bf 0 $ &$ -12.52$ &$ -12.72$ &$\bf 0$ &$-12.43$&$-12.65$&$\bf 0$    \\
Triangulation from satellite im. &$-11.67$&$-11.80$ &$\bf 0 $ &$ -11.53$ &$ -11.83$ &$ 0.76$ &$-11.61$&$-11.93$&$0.5$    \\
Optimal pose 2pt v2 &$-9.85 $& $-10.04$ & $\bf 0.1$  &$-10.85$ &$-10.83$ &$\bf 0.1$ &$-10.36$ &$-10.61$ &$\bf 0.1$ \\
Optimal PnP (Cayley) &$-9.14$&$ -9.45$ &$\bf 3.64 $ &$ -8.38$ &$ -8.74$ &$10.28$ &$-8.42$&$-8.75$&$7.64$    \\
Optimal PnP (Hesch) &$ -11.07$&$ -11.34$ &$ 0.98 $ &$ -11.36$ &$ -11.72$ &$ 0.82$ &$-11.05$ &$-11.36$ &$\bf 0.1$    \\
Unsynch. Rel. pose\textsuperscript{$(\ddagger)$} &$  -10.26$&$ -10.40$ &$\bf 0 $ &$ -8.13$ &$ -8.64$ &$3.84$&$-9.93$&$-10.19$&$0.86$ \\
\bottomrule
\end{tabular}
\caption{A comparison of stability for solvers generated by our proposed resultant-based method, solvers generated using~\cite{larsson2017efficient} and heuristic-based solvers~\cite{DBLP:conf/cvpr/LarssonOAWKP18} on some more minimal problems. Mean and median are computed from $Log_{10}$ of normalized equation residuals. Missing entries are when we failed to extract solutions to all variables. $(\dagger)$: Input polynomials were eliminated using G-J elimination before generating a solver using our resultant method as well as solvers based on~\cite{larsson2017efficient} and the heuristic-based solver~\cite{DBLP:conf/cvpr/LarssonOAWKP18}. $(\ddagger)$: Alternate eigenvalue formulation used for generating the solver based on our proposed method(see Proposition 3.1 in our main paper).}
\label{tbl:stabcomp}
\end{table*}

\begin{figure*}[t]
\centering
\includegraphics[width=0.32\linewidth]{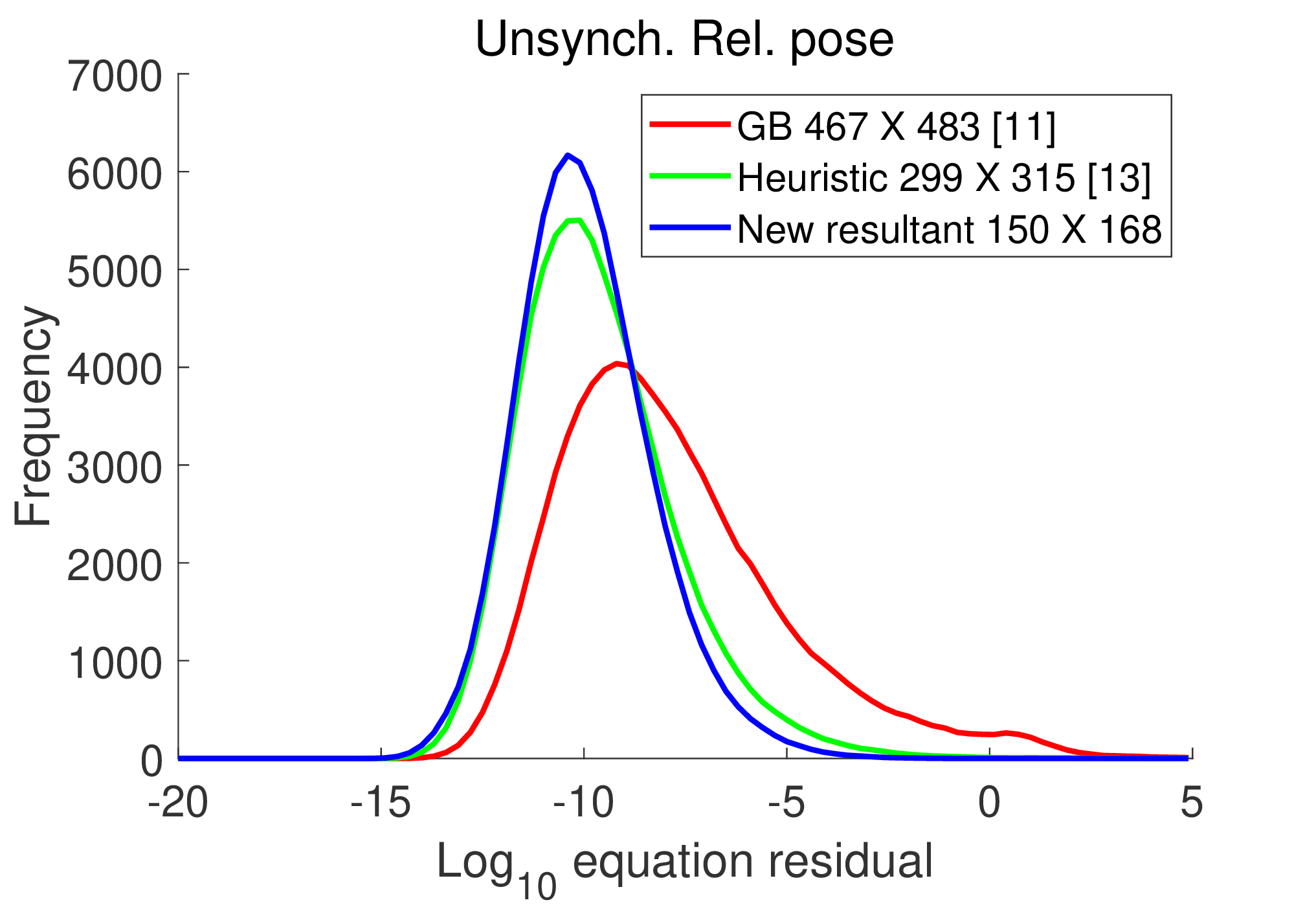}
\includegraphics[width=0.32\linewidth]{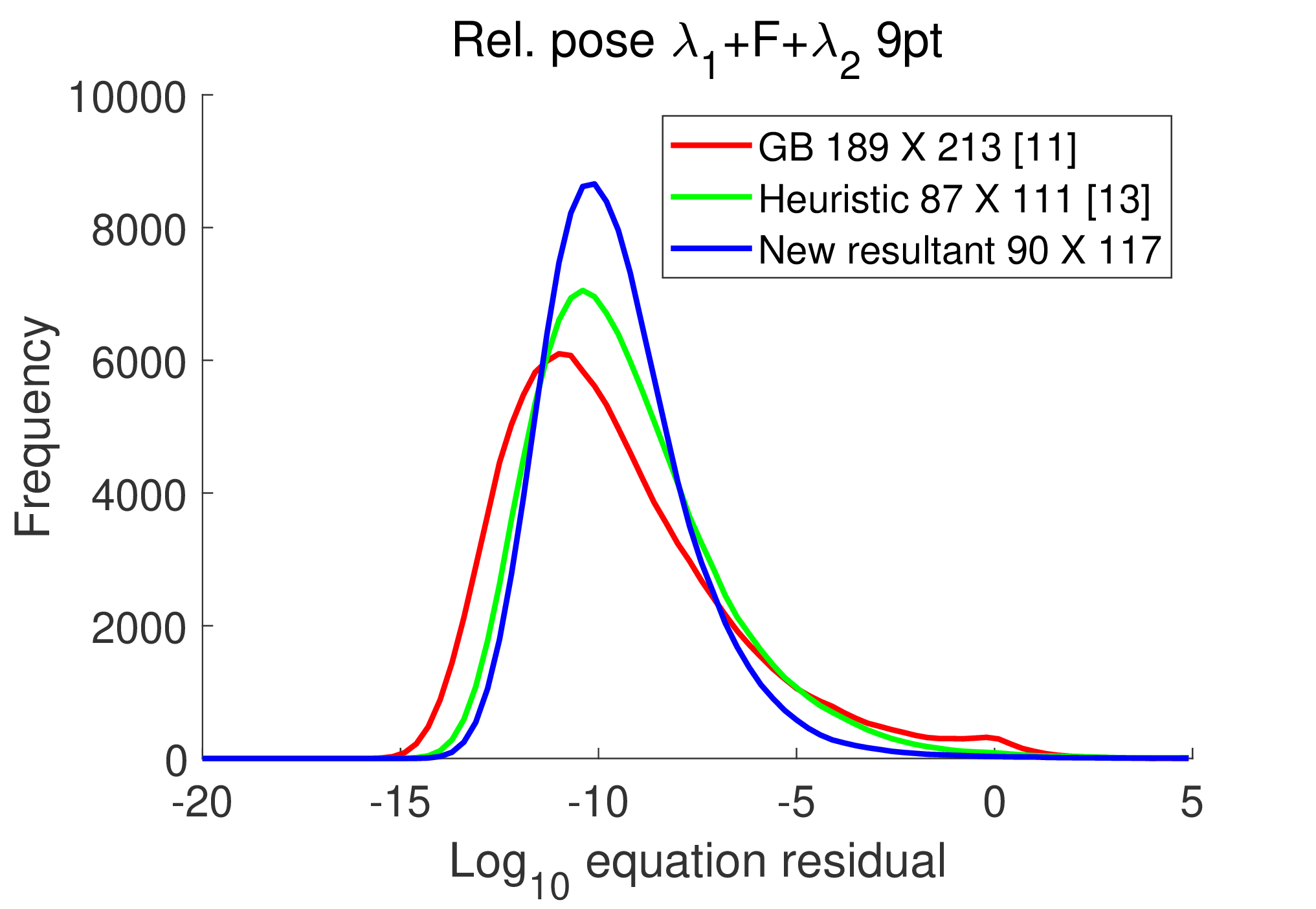}
\includegraphics[width=0.32\linewidth]{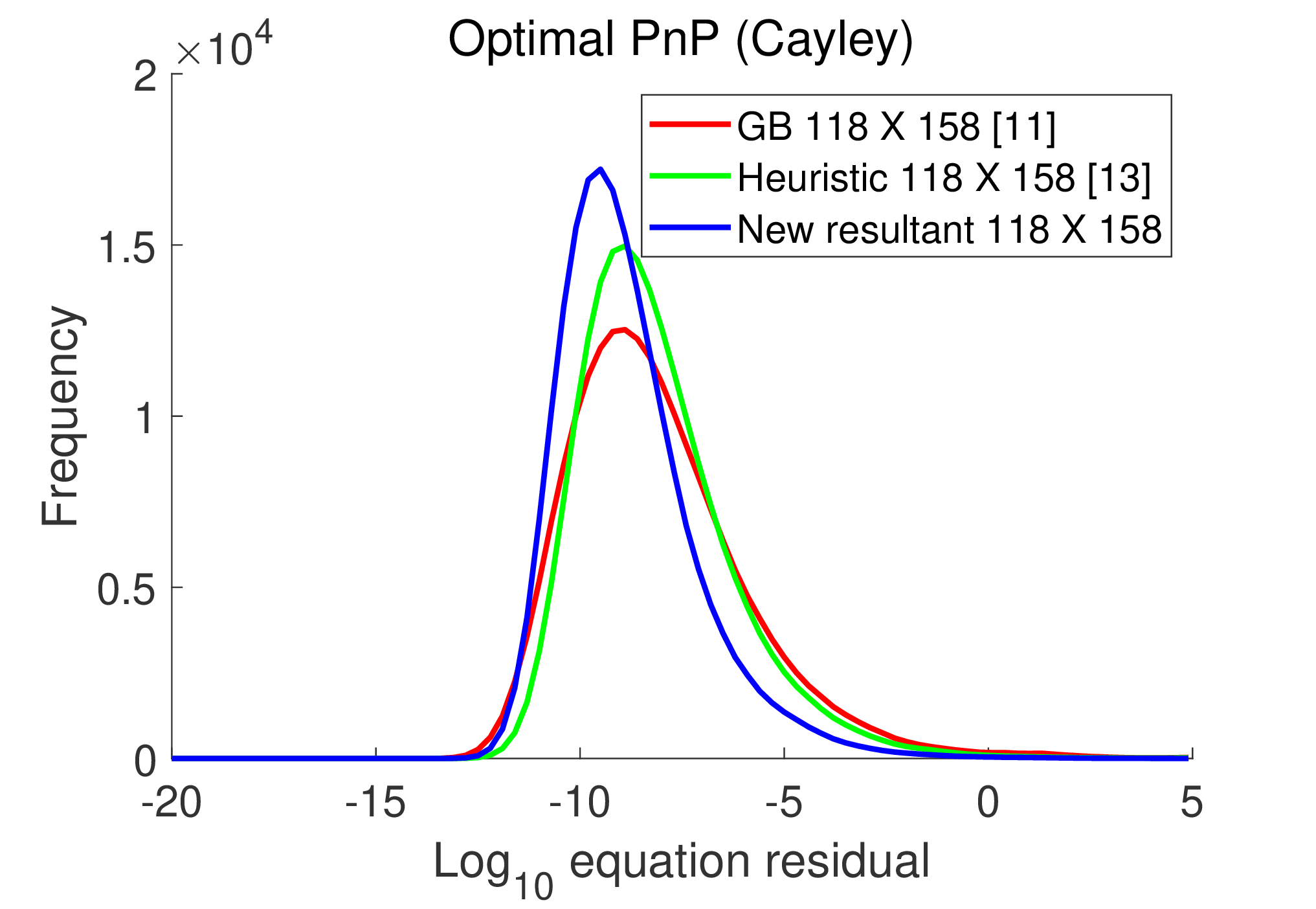}\\
\includegraphics[width=0.32\linewidth]{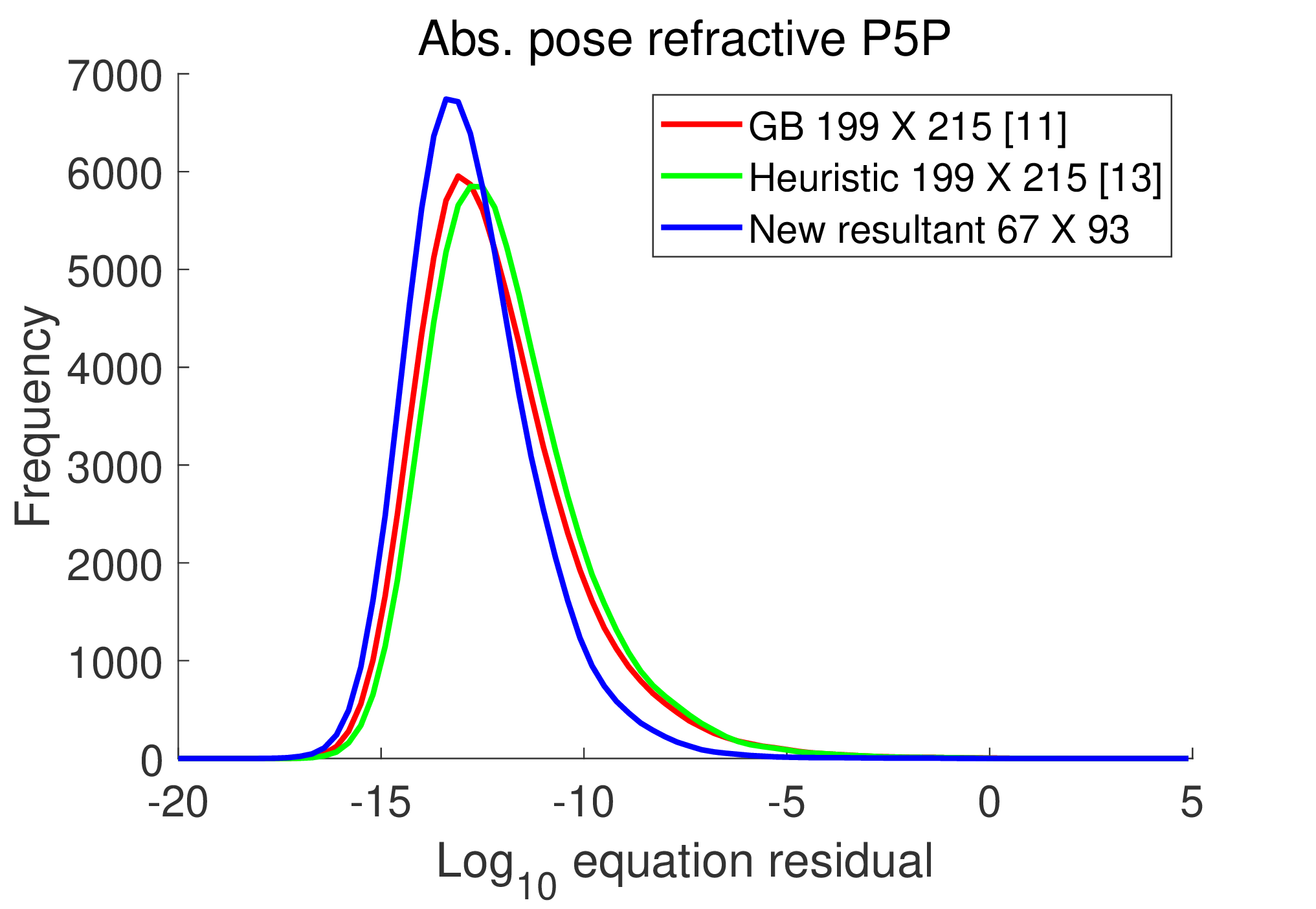}
\includegraphics[width=0.32\linewidth]{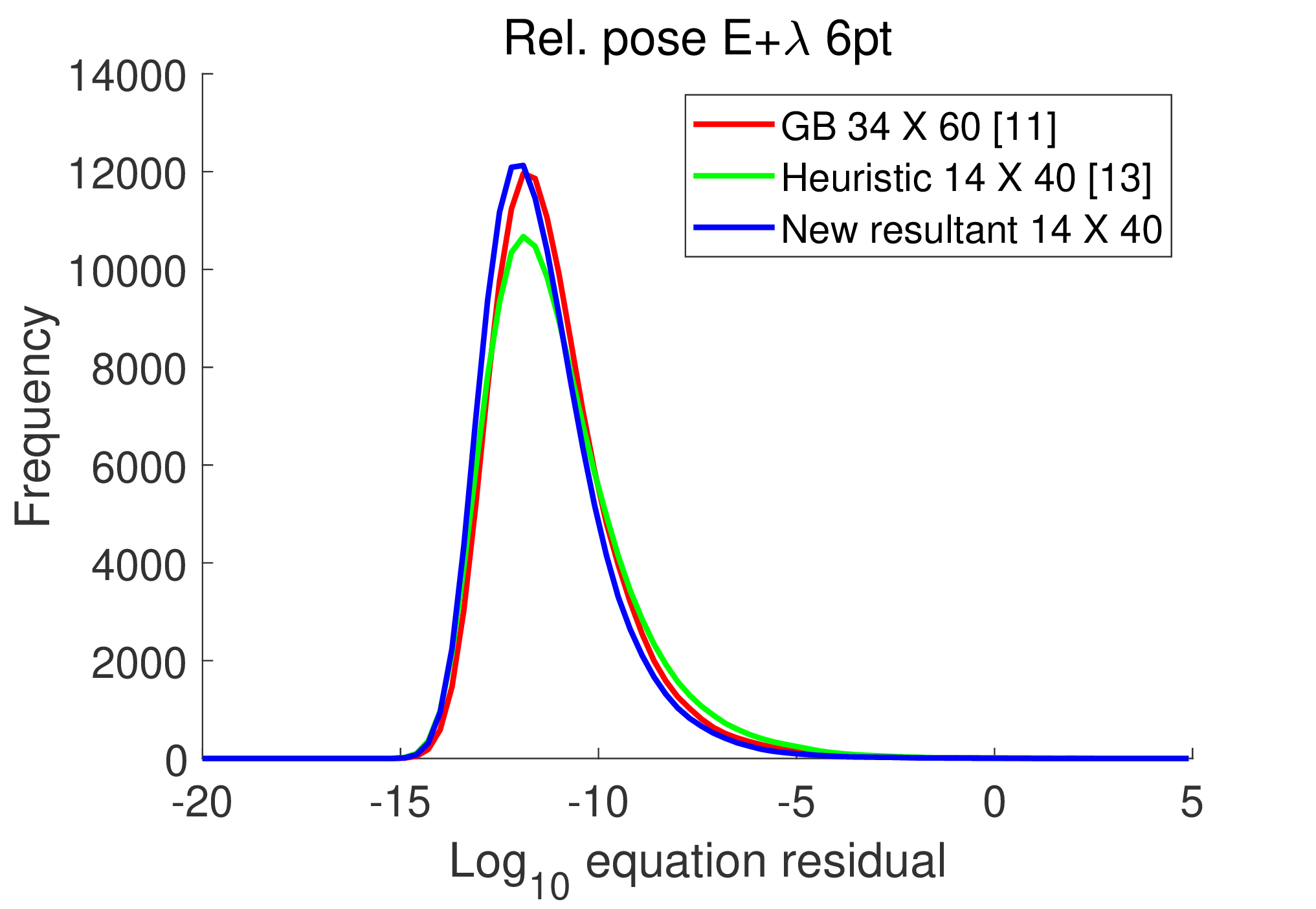}
\includegraphics[width=0.32\linewidth]{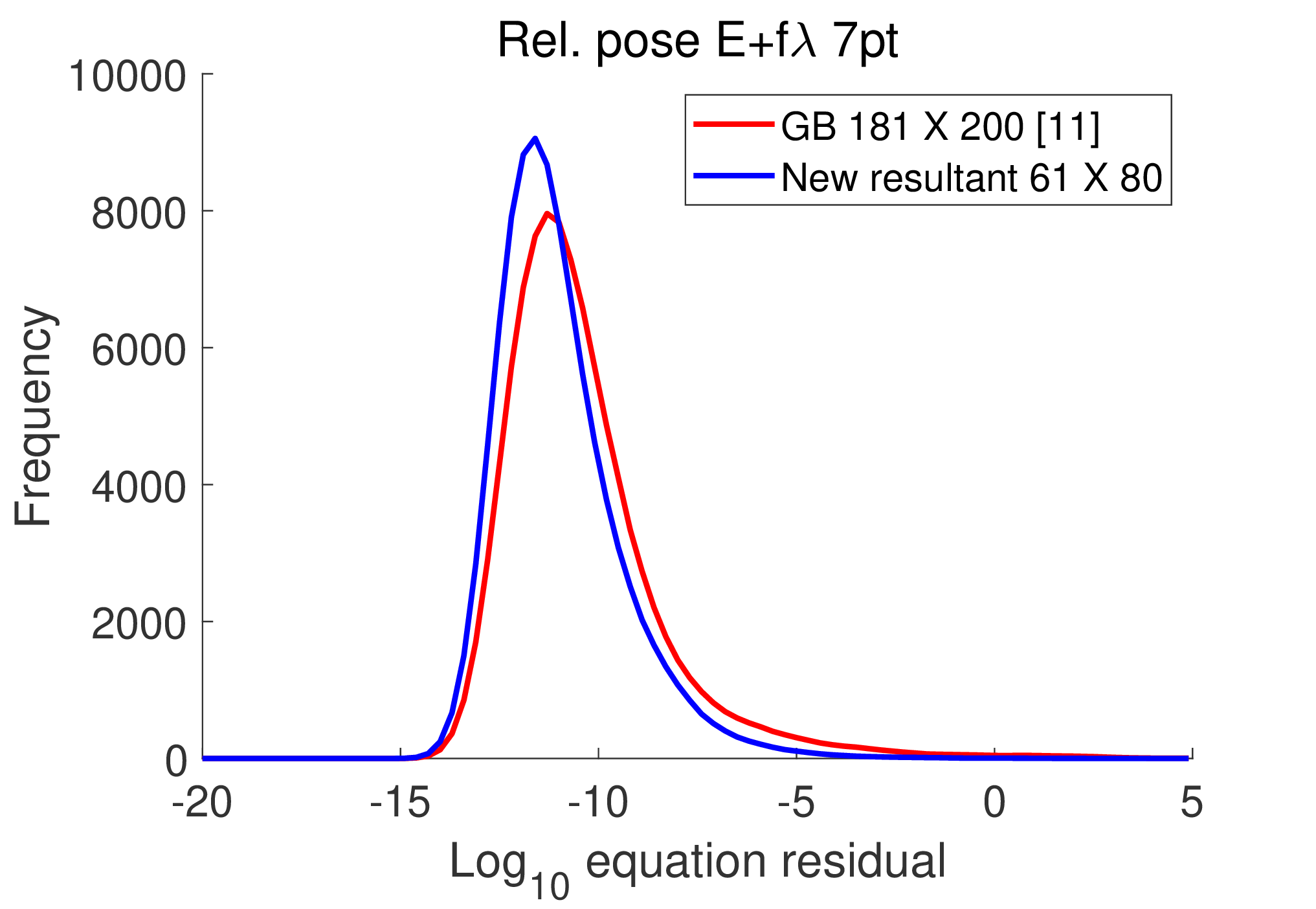}\\
\includegraphics[width=0.32\linewidth]{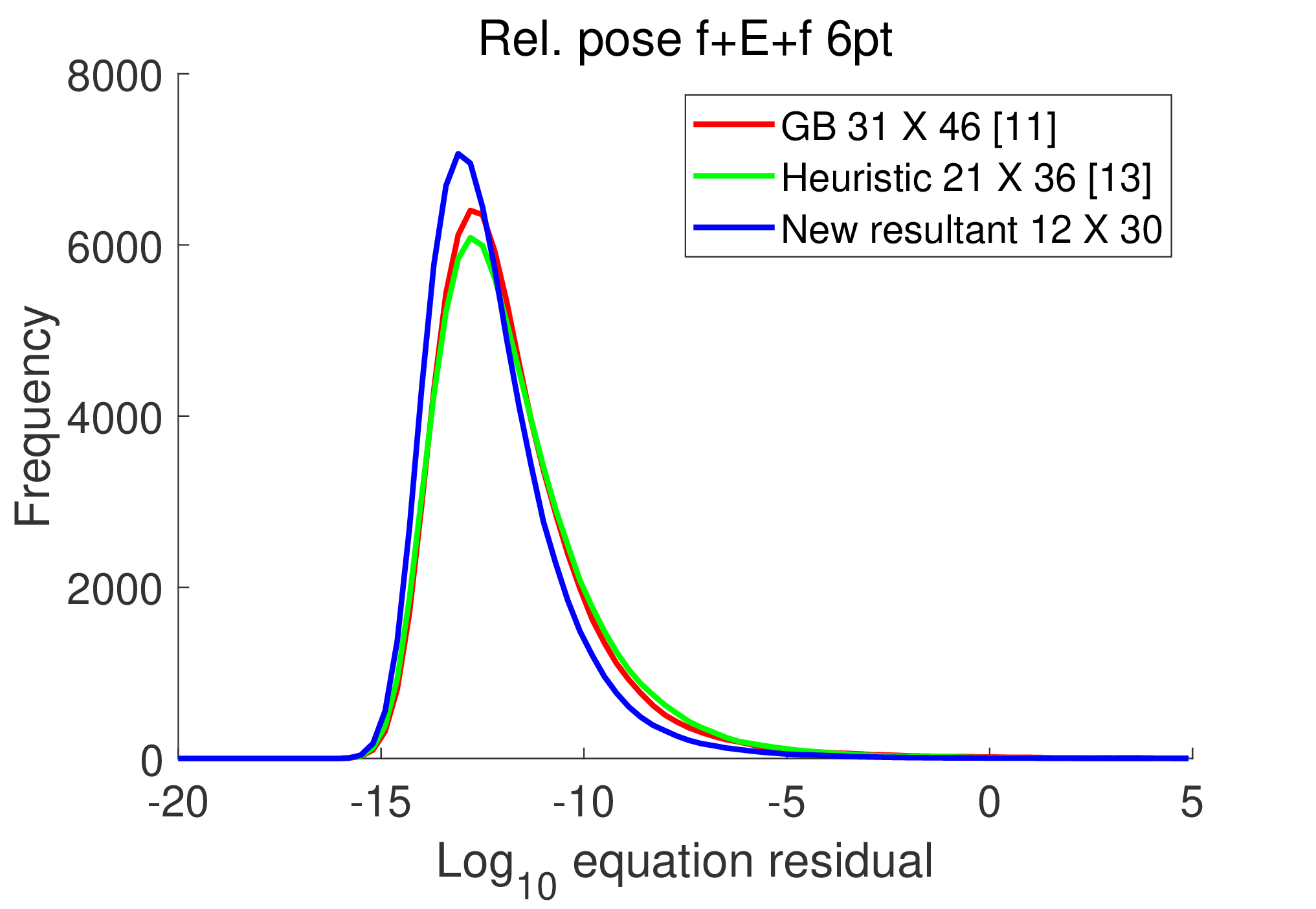}
\includegraphics[width=0.32\linewidth]{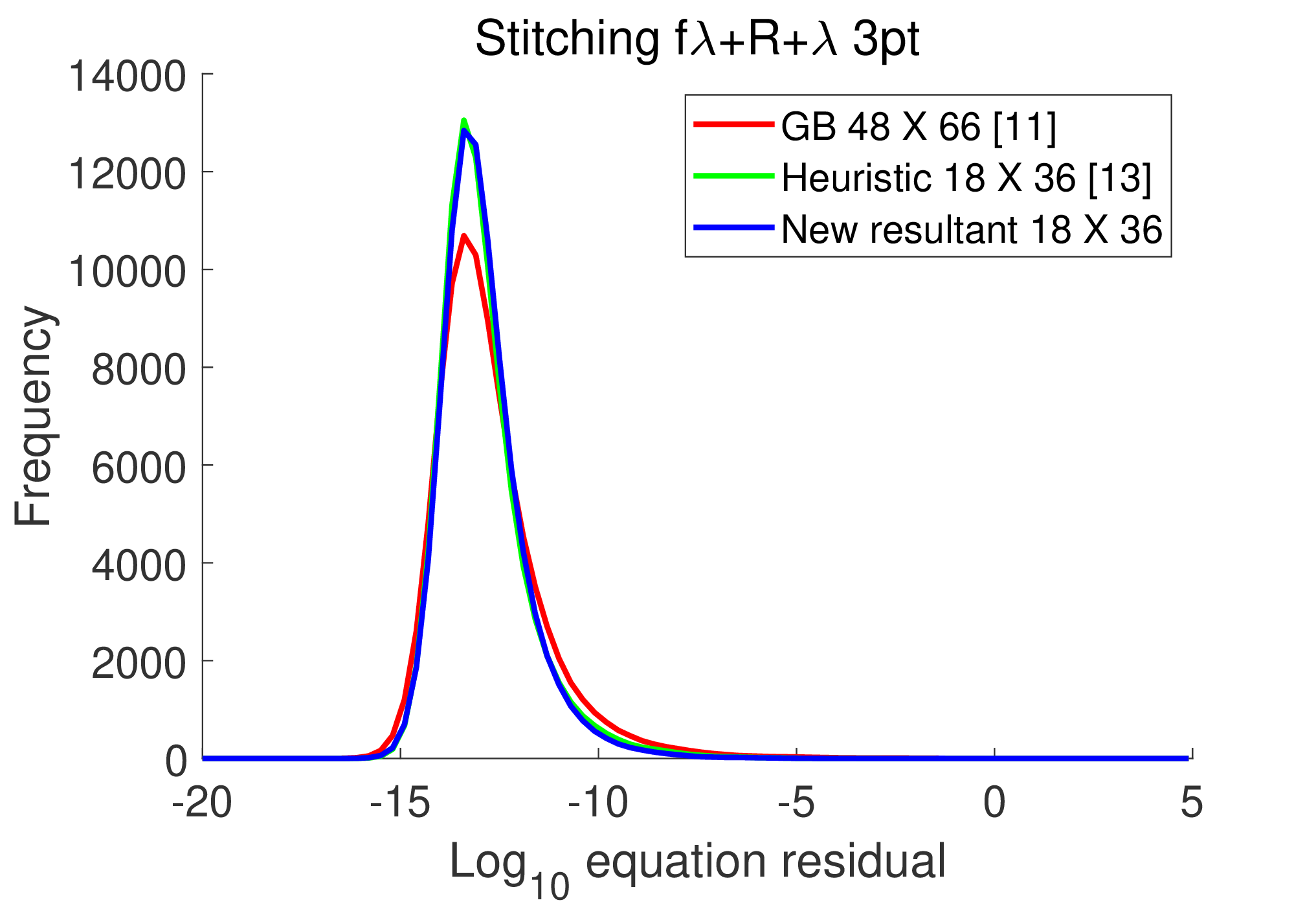}
\includegraphics[width=0.32\linewidth]{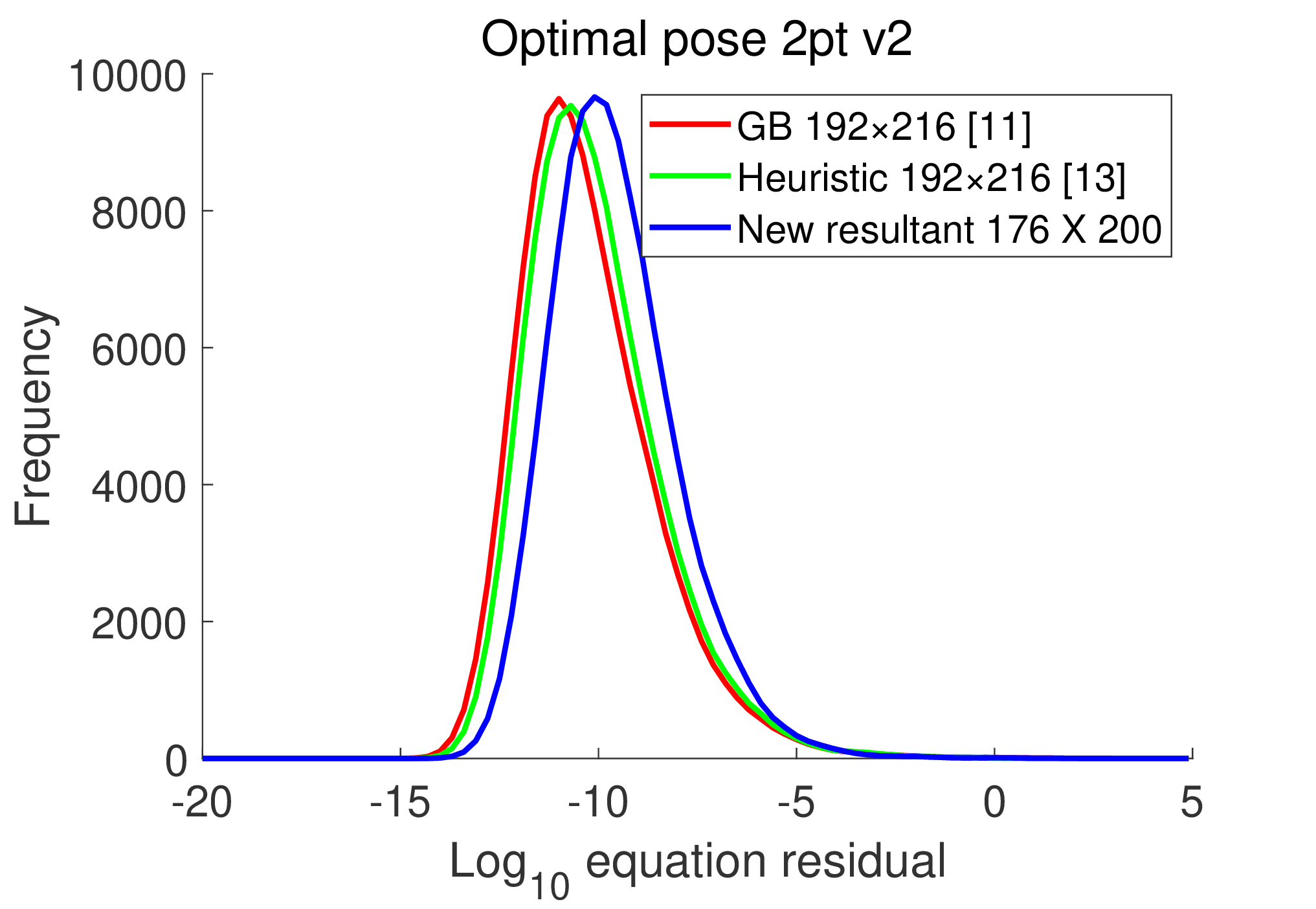}\\
\caption{Histograms of $Log_{10}$ of normalized equation residual error for nine selected minimal problems.
}
\label{fig:histograms}
\end{figure*}

Additionally we provide histograms of residuals in Figure~\ref{fig:histograms} for an interesting set of minimal problems whose stability comparisons have been performed either in Table~\ref{tbl:stabcomp} here or in the Table 2 of our main paper. The residuals have been obtained based on 5K runs on random input data points. We observe from these histograms that our proposed solvers have comparable stability w.r.t. the state-of-the-art solvers based on \gb~\cite{larsson2017efficient} and heuristic-based solvers~\cite{DBLP:conf/cvpr/LarssonOAWKP18}. However an important measure of stability for real world applications is the \% of failures of a minimal solver. Here, we have measured a solver's failure as the number of instances with large values of the equation residual(say above $10^{-3}$) for computed solutions. Using this failure metric, we observe that our proposed resultant-based solvers for the four problems, Unsynch. Rel. pose~\cite{DBLP:journals/corr/AlblKFHSP17}, Rel. pose $\lambda_1$+F+$\lambda_2$ 9pt~\cite{Kukelova-ECCV-2008}, Optimal PnP (Cayley)~\cite{DBLP:conf/bmvc/Nakano15} and Abs. pose refractive P5P~\cite{haner2015absolute} clearly have less failures than the state-of-the-art \gb and heuristic-based solvers. We also note that for four problems from Figure~\ref{fig:histograms}, i.e. Rel. pose $f$+E+$f$ 6pt~\cite{Kukelova-ECCV-2008}, Abs. pose refractive P5P~\cite{haner2015absolute}, Rel. pose E+$f\lambda$ 7pt~\cite{kuang2014minimal} and Optimal pose 2pt v2~\cite{svarm2017city}, our proposed solvers are smaller than the state-of-the-art solvers based on \gb~\cite{larsson2017efficient} and heuristic-based solvers~\cite{DBLP:conf/cvpr/LarssonOAWKP18}. Moreover, for the problem of Unsynch. Rel. pose~\cite{DBLP:journals/corr/AlblKFHSP17}, our proposed solver is significantly smaller than the competitive solvers for the same formulation of the problem.
\begin{figure}[h!]
\centering
\includegraphics[width=0.98\linewidth]{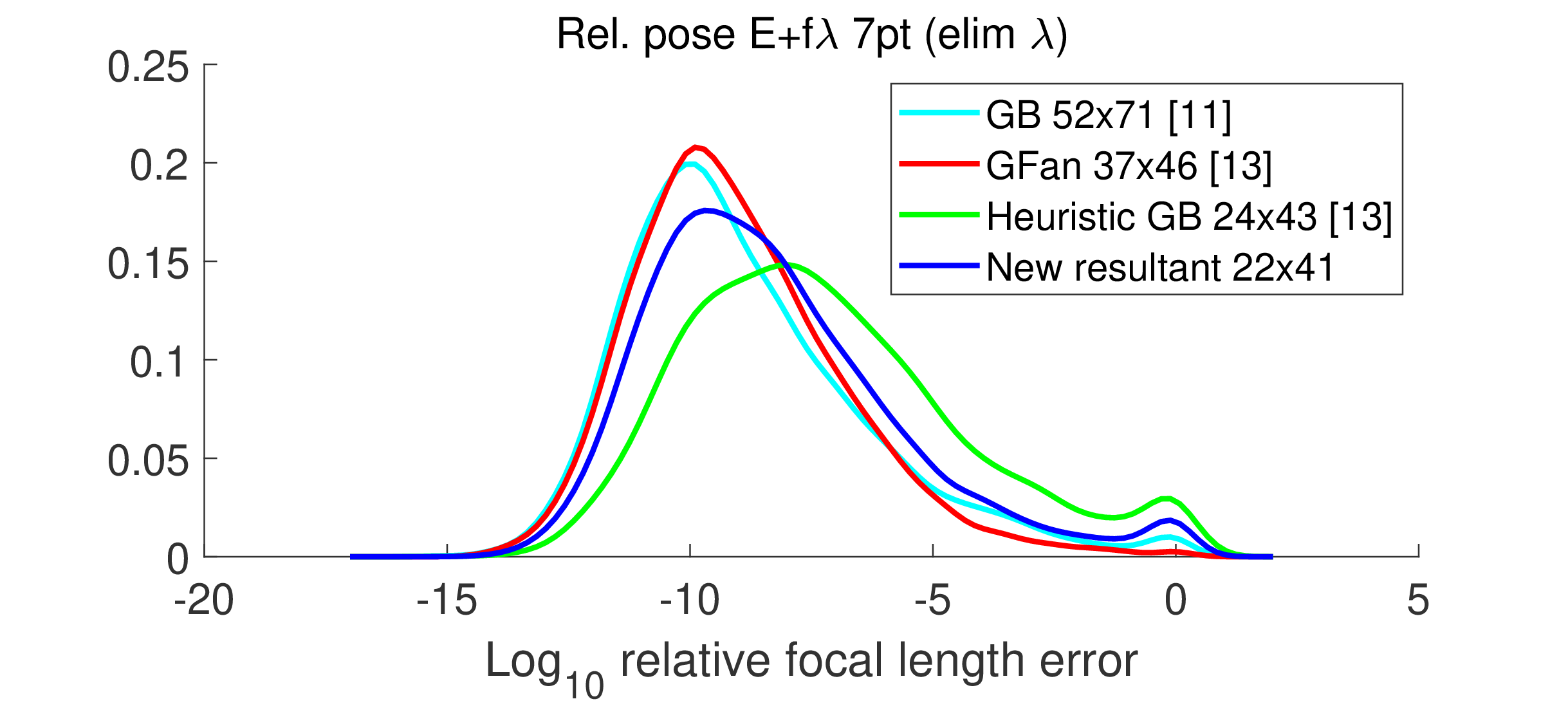}
\caption{Histograms of $Log_{10}$ relative error in focal length for Rel. pose E+$f\lambda$ 7pt (elim$\lambda$) problem 
for 10K randomly generated synthetic scenes. These scenes represent cameras with different radial distortions, poses and focal lengths.
}
\label{fig:7pt}
\end{figure}

\vspace{-0.2cm}
\paragraph{E+$f\lambda$ solver on synthetic scenes:} 
Here we show additional results from the synthetic experiment presented in the main paper. We studied the numerical stability of the new resultant-based solver for the problem of estimating the relative pose of one calibrated and one camera with unknown focal length and radial distortion from 7-point correspondences, i.e.\ the Rel.\ pose E+$f\lambda$ 7pt problem. We considered the formulation ``elim.\ $\lambda$'' proposed in~\cite{DBLP:conf/cvpr/LarssonOAWKP18} that leads to the smallest solvers. We studied the stability on 10K synthetically generated scenes as described in the main paper, see Section 4.1.

Figure~\ref{fig:7pt} shows $Log_{10}$ of the relative error of the focal length obtained by selecting the real root closest to the ground truth $f_{gt}$. The results for radial distortion are in the main paper. All tested solvers provide stable results with only a small number of runs with larger errors. The new resultant-based solver (blue) is not only smaller but also slightly more stable than the heuristic-based solver from~\cite{DBLP:conf/cvpr/LarssonOAWKP18} (green).

\begin{table*}[h!]
\centering
\scalebox{1}{
\begin{tabular}{l c c c | c c c} \toprule
Solver &  \multicolumn{3} {c} {\normalsize Our P4Pfr $28 \times 40$ } &   \multicolumn{3} {c} {\normalsize P4Pfr $40 \times 50$ \cite{larsson2017making}} \\ \cmidrule(r){2-4} \cmidrule(r){5-7}
& avg. & med. & max & avg. & med. & max \\ \midrule
Focal (\%) & 0.080 & 0.063 & 0.266 & 0.08 & 0.07 & 0.29     \\
Distortion  (\%)  & 0.522 & 0.453 & 1.651 & 0.51 & 0.45 & 1.85      \\
Rotation  (degree)  &0.031 & 0.029 & 0.062 &0.03 & 0.03 & 0.10    \\
Translation (\%) & 0.066 & 0.051 & 0.210 & 0.07 & 0.07 & 0.26   \\ \bottomrule
\end{tabular}}
~
\caption{Errors for the real Rotunda dataset. The errors are relative to the ground truth for all except rotation which is shown in degrees. The results for the P4Pfr solver ($40 \times 50$) \cite{larsson2017making} are taken from~\cite{larsson2017making}}
\label{tbl:real_data}
\end{table*}

\paragraph{P4Pfr solver on real images:} \label{sec:real_images}
\noindent Here we show additional statistics for the real experiment presented in our main paper where we evaluated our proposed solver for the problem of estimating the absolute pose of a camera with unknown focal length and radial distortion from four 2D-to-3D point correspondences, i.e. the P4Pfr solver. We consider the \textit{Rotunda} dataset, which was proposed in \cite{kukelova2015efficient} and in~\cite{larsson2017making} it was used for evaluating P4Pfr solvers.  This dataset consists of 62 images captured by a GoPro Hero4 camera with significant radial distortion. The Rotunda reconstruction contains 170994 3D points and the average reprojection error was 1.4694 pixels over 549478 image points.  We used the 3D model to estimate the pose of each image using our new P4Pfr resultant-based solver ($28\times 40$) in a RANSAC framework. Similar to \cite{larsson2017making}, we used the camera and distortion parameters obtained from \cite{realitycapture} as ground truth for the experiment.

In Table~\ref{tbl:real_data} we present the errors for the focal length, radial distortion, and the camera pose obtained using our proposed solver and for the sake of comparison we also list the errors, which were reported in ~\cite{larsson2017making}, where the P4Pfr (40x50) solver was tested on the same dataset. Overall the errors are quite small, \eg most of the focal lengths are within $0.1\%$ of the ground truth and almost all rotation errors are less than $0.1$ degrees, which shows that our new solver as well as the original solver work well for real data. The results of both solvers are very similar. However, we do take note that the slightly different values of errors are mainly due to RANSAC's random nature.

{\small
\bibliographystyle{ieee}
\bibliography{supp}
}